**Do large language models resemble humans in language use?**

Zhenguang G. Cai[1,2], Xufeng Duan[1], David A. Haslett[1], Shuqi Wang[1], Martin J. Pickering[3]
[1] Department of Linguistics and Modern Languages, The Chinese University of Hong Kong
[2] Brain and Mind Institute, The Chinese University of Hong Kong
[3] Department of Psychology, University of Edinburgh

*Zhenguang G. Cai
Email: zhenguangcai@cuhk.edu.hk

**Author Contributions:** ZGC conceived the study; ZGC, DAH and XD designed the experiments; MJP provided critical comments on study designs; SW, DAH, XD, and ZGC constructed experimental stimuli; ZGC, DAH, SW, and XD preregistered the experiments; XD collected the data; XD, SW, DAH and ZGC coded the data; ZGC, XD, and DAH analyzed the data; DAH and ZGC wrote the first draft. All authors contributed to the write-up.

**Competing Interest Statement:** The authors declare no competing interest.

**Author Notes:**
This is a revised version of an earlier preprint entitled "Does ChatGPT resemble humans in language use?". The title was updated to reflect the fact that we have now also included data from an open-source language model Vicuna. The ordering of authorship was also updated to reflect contribution of the authors.

**Significance Statement**

Large language models like ChatGPT and Vicuna have the potential to reshape society, yet their resemblance to human language use remains uncertain. We conducted 12 preregistered psycholinguistic tests, spanning from phonetics to syntax to interpersonal communication, on these models. ChatGPT exhibited human-like responses in 10 experiments, with Vicuna showing similar patterns in 7. This robust evidence suggests that ChatGPT, and to a lesser extent Vicuna, mirror fundamental aspects of human language processing. Our findings underscore the profound implications of large language models, showcasing their capacity to manifest human-like linguistic behaviors. Moreover, they signify a pivotal step in bridging the gap between human and machine language, promising deeper insights into both linguistic cognition and artificial intelligence's potential societal impact.

**Abstract**

Large language models (LLMs) such as ChatGPT and Vicuna have shown remarkable capacities in comprehending and producing language. However, their internal workings remain a black box, and it is unclear whether LLMs and chatbots can develop humanlike characteristics in language use. Cognitive scientists have devised many experiments that probe, and have made great progress in explaining, how people comprehend and produce language. We subjected ChatGPT and Vicuna to 12 of these experiments ranging from sounds to dialogue, pre-registered and with 1,000 runs (i.e., iterations) per experiment. ChatGPT and Vicuna replicated the human pattern of language use in 10 and 7 out of the 12 experiments, respectively. The models associated unfamiliar words with different meanings depending on their forms, continued to access recently encountered meanings of ambiguous words, reused recent sentence structures, attributed causality as a function of verb semantics, and accessed different meanings and retrieved different words depending on an interlocutor's identity. In addition, ChatGPT, but not Vicuna, nonliterally interpreted implausible sentences that were likely to have been corrupted by noise, drew reasonable inferences, and overlooked semantic fallacies in a sentence. Finally, unlike humans, neither model preferred using shorter words to convey less informative content, nor did they use context to resolve syntactic ambiguities. We discuss how these convergences and divergences may result from the transformer architecture. Overall, these experiments demonstrate that LLMs such as ChatGPT (and Vicuna to a lesser extent) are humanlike in many aspects of human language processing.

Large language models (LLMs, e.g., BERT, GPT models, and related chatbots such as ChatGPT) have demonstrated remarkable capacities in producing humanlike language (e.g., Brown et al., 2020; Devlin et al., 2019; Ouyang et al., 2022; Radford et al., 2018, 2019). They can supply missing words in sentences, interpret word meanings in context, and judge the grammaticality of sentences (e.g., Gulordava et al., 2018; Liu et al., 2019; Mahowald, 2023; Marvin & Linzen, 2018; Peters et al., 2018; Qiu et al., 2024; Tenney et al., 2019). The formal linguistic competence apparent in LLMs has led to debates over whether they can serve as cognitive models of the human language (see Mahowald et al., 2023). On the one hand, Chomsky argued that humans are endowed with an innate universal grammar (e.g., Chomsky, 2000), and he and colleagues maintain that this "genetically installed 'operating system'... is completely different from that of a machine learning program" (Chomsky et al., 2023, para. 6) such as ChatGPT, which is simply "a lumbering statistical engine for pattern matching" (para. 5). More optimistic researchers, however, argue that deep neural networks suffice to learn syntactic structure (Piantadosi, 2023), as evidenced by the fact that LLMs abide by complex grammatical rules (e.g., Goldberg, 2019; Linzen & Baroni, 2021; McCoy et al., 2019).

This debate emphasizes grammar, but regularities in language range from phonology to pragmatics. For example, people associate different sounds with different referents (e.g., Köhler, 1929), automatically reinterpret implausible sentences (e.g., Gibson et al., 2013), and expect demographically appropriate content from speakers (e.g., Van Berkum et al., 2008). Do LLMs share these regularities in language use? Piantadosi (2023) pointed out that LLMs integrate syntax and semantics (i.e., all aspects of usage are represented in a single vector space), so other humanlike regularities in language use might emerge along with grammaticality and coherence. Supporting the hypothesis that humans and LLMs have underlying similarities, representations from intermediate layers of LLMs (rather than initial layers, which are not contextualized, or output layers) are highly predictive of activation in language-selective brain regions when humans and LLMs process the same passage of text (Caucheteux & King, 2022; Schrimpf et al., 2021). Whether LLMs have developed humanlike regularities will not only gauge the success of computational research into natural language processing, but will be important for cognitive scientists. For instance, as with universal grammar, it is a matter of debate whether regularities in language processing arise from innate constraints or statistical learning. Given that LLMs do not have built-in linguistic rules, they allow us to test which regularities in language use can be recovered from patterns in language—at least in principle, with an abundance of training data.

We therefore subjected two LLMs—ChatGPT, from OpenAI (2022), and Vicuna (with 13B parameters), from the Large Model Systems Organization (Chiang et al., 2023)—to a battery of psycholinguistic tests, in 12 preregistered experiments per LLM. These experiments span a range of linguistic levels from sounds to discourse, with two experiments per level. In each experiment, each item was presented to each LLM 1000 times. Table 1 summarizes the experiments and their key findings; the preregistrations, data, and analytical codes are available at osf.io/vu2h3/ (ChatGPT) and osf.io/sygku/ (Vicuna).

**Table 1.** An overview of the tasks, manipulations and key findings in the 12 pre-registered experiments.

| **Sounds: sound-shape association** |
|---|
| When guessing whether a nonword refers to a round or spiky shape, both LLMs gave more round judgements for nonwords judged by humans to sound round (e.g., *maluma*) than spiky (e.g., *takete*). |
| **Sounds: sound-gender association** <br> When completing a sentence preamble containing a novel name (e.g., *Although Pelcrad/Pelcra was sick...*), both LLMs used more feminine pronouns (i.e., *she/her/hers*) to refer to the novel name ending in a vowel than a consonant. |
| **Words: word length and predictivity** <br> Humans more often choose the shorter of two words with very similar meanings (e.g., *mathematics* vs *math*) when completing a sentence that is predictive of the word's meaning (e.g., *Susan was very bad at algebra, so she hated…*) than when completing a neutral sentence (e.g., *Susan introduced herself to me as someone who hated…*). Neither LLM exhibited this tendency. |
| **Words: word meaning priming** <br> After an ambiguous cue word (e.g., *post*), both LLMs supplied an associate of a particular meaning of that word (e.g., *position*, rather than *mail*) more often when they had earlier read a sentence that used that meaning of the ambiguous word (e.g., *The man accepted the post in the accountancy firm*) than when they had read a sentence containing a synonym (e.g., *The man accepted the job in the accountancy firm*) or when they had not read a sentence containing that meaning. |
| **Syntax: structural priming** <br> When completing a prime preamble (e.g., *The racing driver gave the torn overall ...*) and then a target preamble (e.g., *The patient showed . . .*), both ChatGPT and Vicuna tended to use the same syntactic structure in the prime and the target. For example, they were more likely to compete a target preamble into a prepositional-objective (PO) dative sentence (e.g., *The patient showed his hand to the nurse*) after completing a prime preamble into a PO sentence (e.g., *The racing driver gave the torn overall to his mechanic*) than after completing a prime preamble into a double-object (DO) dative sentence (e.g., *The racing driver gave the helpful mechanic a wrench*). In both models, the priming effect was enhanced when the prime and the target had the same verb (*showed*) compared to different verbs (*showed* vs *gave*). |
| **Syntax: syntactic ambiguity resolution** <br> Humans more often interpret the syntactically ambiguous phrase such as *with the rifle* in *The hunter shot the dangerous poacher with the rifle* as modifying the noun (i.e., the poacher had the rifle) than the verb (i.e., the hunter used the rifle) when the discourse has introduced multiple potential referents than a single referent for *the dangerous poacher* (e.g., *There was a hunter and two poachers / a poacher*). Neither LLMs exhibited this tendency. |
| **Meaning: implausible sentence interpretation** <br> After reading an implausible sentence in a DO structure (e.g., *The mother gave the candle the daughter*) or in a PO structure (e.g., *The mother gave the daughter to the candle*), ChatGPT, but not Vicuna, was more likely to nonliterally interpret the implausible DO than PO sentence (e.g., treating the daughter as the recipient of the candle). |
| **Meaning: semantic illusion** <br> ChatGPT, but not Vicuna, noticed fewer errors when sentences contained incongruent words that were semantically close to congruent words (e.g., *Snoopy is the black and white cat in what famous Charles Schulz comic strip?*) than when they were semantically distant (e.g., |
| *Snoopy is the black and white mouse in what famous Charles Schulz comic strip?*; in fact, Snoopy is a dog). |
| **Discourse: implicit causality** <br> When completing a sentence preamble (*Gary scared/feared Anna because …*) into a full sentence, both LLMs were more likely to attribute the causality of the preamble evenet to the object rather than the subject (e.g., *Anna* rather than *Gary*) when the preamble had a *stimulus-experiencer* verb with the subject serving as the stimulus and the object serving as the experiencer (e.g., *Gary scared Anna because he was violent*) than when the preamble had an *experiencer-stimulus* verb (e.g., *Gary feared Anna because she was violent*). |
| **Discourse: drawing inferences** <br> ChatGPT, but not Vicuna, was more likely to make inferences that connect two pieces of information (e.g., *Sharon stepped on glass. She cried out for help.*) than to make inferences that elaborate on a single piece of information (e.g., *Sharon stepped on glass. She was looking for a watch.*), in response to a question (e.g., *Did she cut her foot?*). |
| **Interlocutor sensitivity: word meaning access** <br> When supplying an associate to a word with different meanings in different dialects (e.g., *bonnet* meaning "car-part" in British English but "hat" in American English), both LLMs were more likely to access the American English meaning when the interlocutor self-identified as an American English speaker than as a British English speaker. |
| **Interlocutor sensitivity: lexical retrieval** <br> When asked to supply a word/phrase according to a provided definition (e.g., *a housing unit common in big cities that occupies part of a single level in a building block*), both LLMs were more likely to retrieve an American expression instead of a British one (e.g., *apartment* vs. *flat*) when the interlocutor identified him/herself as an American English speaker than as a British English speaker. |

## RESULTS

*Sounds: sound-shape association*

People tend to associate certain sounds with certain shapes. They assume, for instance, that a novel word such as *takete* or *kiki* refers to a spiky object, whereas a novel word such as *maluma* or *bouba* refers to a round object (Ćwiek et al., 2022; Köhler, 1929). We presented ChatGPT (9 Jan 2023 version; osf.io/6wxp3) and Vicuna (osf.io/6zhvx) with 10 novel words deemed round-sounding by human participants and 10 novel words deemed spiky-sounding, adapted from Sidhu and Pexman (2017), and asked ChatGPT/Vicuna to decide whether each word refers to a spiky shape or a round shape.

Both LLMs assigned round-sounding novel words to round shapes more often than they assigned spiky-sounding novel words to round shapes (ChatGPT: 0.79 vs. 0.49, $\beta = 2.02$, $SE = 0.34$, $z = 5.87$, $p < .001$; Vicuna: 0.38 vs. 0.32, $\beta = 0.27$, $SE = 0.11$, $z = 2.34$, $p = .019$; see Fig 1 top left). We note ChatGPT and/or Vicuna might have been exposed to the abstract and/or full paper of the human study that our LLM experiments replicated (e.g., Sidhu & Pexman, 2017, for the current experiments). So in all experiments in this study, we performed non-registered analyses where we excluded experimental items that were mentioned in the relevant studies (see Supplemental Information for details). For the current experiments, we observed the same pattern of results even after excluding exemplar stimuli *maluma* and *takete* (which were used as examples in Sidhu & Pexman, 2017). In the following experiments, the same pattern of results was obtained after excluding exemplar stimuli, unless otherwise mentioned (and see the Supplement Information for the analyses).

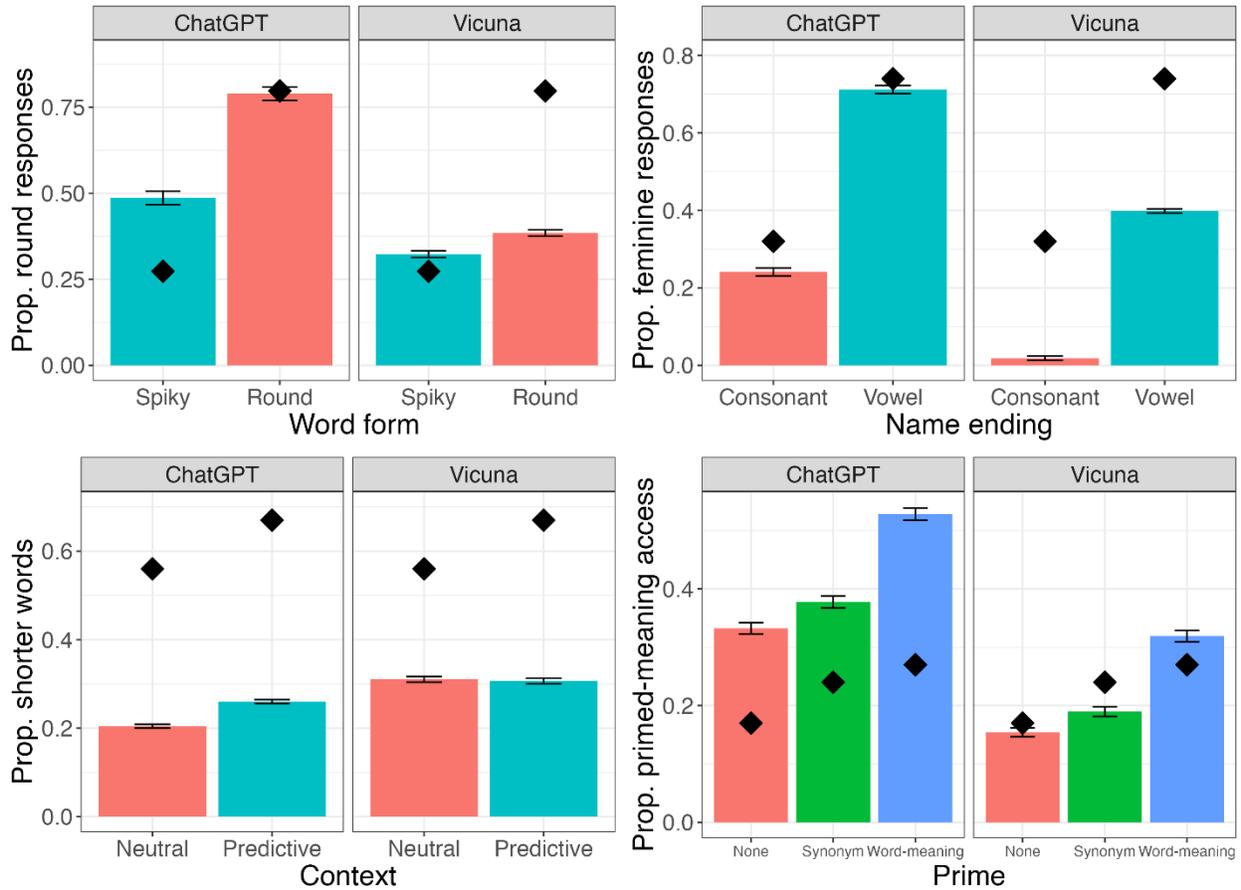

**Fig 1**. Results of sound-shape associations (top left; human conditional means, shown as diamonds, are from Sidhu & Pexman, 2017), sound-gender associations (top right; human conditional means are from Cassidy et al., 1999), word length and predictivity (bottom left; human conditional means are from Mahowald et al., 2013), and word meaning priming (bottom right; human conditional means are from Rodd et al., 2013). Error bars stand for 95% confidence intervals. Note that in some cases, materials have been modified, so the human means are for comparison only.

### *Sounds: sound-gender association*

People can guess at above-chance rates whether an unfamiliar name refers to a man or a woman based on how it sounds (Cai & Zhao, 2019; Cassidy et al., 1999; Cutler et al., 1990). In English, for example, women's names end in vowels more often than men's names do (Cutler et al., 1990). We asked ChatGPT (13 Feb 2023 version; osf.io/7yrf8) and Vicuna (osf.io/g67eb) to complete 16 preambles containing a consonant-ending name or vowel-ending novel name (e.g., 1a-b; adapted from Cassidy et al., 1999).

    1a. Consonant-ending name: *Although Pelcrad was sick...*
    1b. Vowel-ending name: *Although Pelcra was sick...*

Both LLMs were more likely to use a feminine pronoun (*she/her/hers*; e.g., *Although Pelcra was sick,* she *refused to stay in bed and insisted on completing all her tasks for the day*) to refer to vowel-ending names than to consonant-ending names (ChatGPT: 0.71 vs. 0.25, $\beta = 4.33$, $SE = 1.24$, $z = 3.50$, $p < .001$; Vicuna: 0.40 vs. 0.02, $\beta = 5.77$, $SE = 1.23$, $z = 4.70$, $p < .001$; see Fig 1 top right). These findings suggest that the models have learned sound-gender associations in personal names and use these cues to infer gender, as humans do.

*Words: word length and predictivity*

Corporal evidence suggests that words which carry less information tend to be shorter, making communication more efficient (e.g., Piantadosi et al., 2011). In support of this hypothesis, Mahowald et al. (2013) showed that, when asked to choose between a shorter and a longer word of nearly identical meanings (e.g., *math* and *mathematics*), participants more often chose the shorter word to complete a sentence preamble that was predictive of the meaning of the to-be-chosen word (i.e., the word is less informative; e.g., 2a) than to complete a neutral sentence preamble (e.g., 2b). If ChatGPT is sensitive to the relationship between length and informativity, it should opt for shorter words when they are predictable from the context more often than when they are unpredictable.

> 2a. Predictive context: *Susan was very bad at algebra, so she hated... 1. math  2. mathematics*
> 2b. Neutral context: *Susan introduced herself to me as someone who loved... 1. math   2. Mathematics*

We replicated the behavioural study in Mahowald et al. (2013) using ChatGPT (15 Dec 2022 version; osf.io/n645c) and Vicuna (osf.io/8tkf3). Neither LLM was significantly more likely to choose shorter words following predictive preambles than neutral preambles (ChatGPT: 0.26 vs. 0.20, $\beta = 0.35$, $SE = 0.21$, $z = 1.64$, $p = .101$; Vicuna: 0.31 vs. 0.31, $\beta = -0.15$, $SE = 0.20$, $z = -0.77$, $p = .444$; see Fig 1 bottom left). Given that the current experiment was a well-powered and that the effect has been observed repeatedly in humans (e.g., Mahowald et al., 2013; Dunn & Cai, 2023), the lack of the effect here suggest that ChatGPT and Vicuna behave differently from humans in that they do not tend to use a shorter word in a more predictive context.

*Words: word meaning priming*

Words often have multiple meanings, such as *post* referring to mail versus a job. In what is known as word meaning priming (e.g., Rodd et al., 2013), people tend to access the more recently encountered meaning of an ambiguous word. For example, participants more often supplied associates related to the job meaning of *post* (e.g., *work*) if they had recently read a sentence using that meaning (e.g., 3a) than if they had recently read a sentence using a synonym (e.g., 3b) or if they had not read such a sentence (e.g., 3c). Following Rodd et al. (2013), we first presented ChatGPT (9 Jan 2022 version; osf.io/ym7hg) and Vicuna (osf.io/z4ws8) with a set of 44 sentences (adapted from Rodd et al., 2013), including 13 word-meaning primes, 13 synonym primes, and 18 filler sentences; afterwards, we presented them with 39 ambiguous cue words (e.g., *post*) and asked the models to provide an associate, with 13 words per condition, and we measured the proportion of associates related to the primed meaning (e.g., *work*).

3a. Word-meaning prime: *The man accepted the post in the accountancy firm.*
3b. Synonym prime: *The man accepted the job in the accountancy firm.*
3c. No prime: [NO SENTENCE]

Neither LLMs produced significantly more associates related to the primed meaning in the synonym condition than the no-prime condition (ChatGPT: 0.38 vs. 0.33, $β = 0.36$, $SE = 0.19$, $z = 1.90$, $p = .057$; Vicuna: 0.19 vs. 0.15, $β = 0.39$, $SE = 0.28$, $z = 1.40$, $p = .162$; see Fig 1 bottom right). Crucially, both models produced more associates related to the primed meaning in the word-meaning condition than in the no-prime condition (ChatGPT: 0.53 vs. 0.33, $β = 2.47$, $SE = 0.30$, $z = 8.20$, $p < .001$; Vicuna: 0.32 vs. 0.15, $β = 3.33$, $SE = 0.50$, $z = 6.70$, $p < .001$) and also than in the synonym condition (ChatGPT: 0.53 vs. 0.38, $β = 2.14$, $SE = 0.32$, $z = 6.71$, $p < .001$; Vicuna: 0.32 vs. 0.19, $β = 2.86$, $SE = 0.48$, $z = 5.91$, $p < .001$). These finding suggest that both LLMs are susceptible to word-meaning priming.

*Syntax: structural priming*

People tend to repeat a syntactic structure that they have recently encountered, a phenomenon known as structural priming (e.g., Bock, 1986; Pickering & Branigan, 1998). For instance, Pickering & Branigan (1998) had participants first complete a prime preamble that was designed to induce a completion of either a double-object (DO) dative structure (e.g., *The racing driver gave the helpful mechanic a wrench* as a completion for 4a/c) or a prepositional-object (PO) dative structure (PO, e.g., *The racing driver gave the torn overall to his mechanic* as a completion for 4b/d) and then complete a target preamble that could be continued as either a DO or a PO (e.g., 4e). Participants tended to complete a target preamble using the same structure that they used in completing a prime preamble, and the priming effect was larger when the target had the same verb as the prime (e.g., 4c/d) than when it had a different verb from the prime (e.g., 4a/b) (i.e., a lexical boost to structural priming). Following Pickering & Branigan (1998, Experiment 1), we presented ChatGPT (9 Jan 2023 version and 13 Feb 2023 version; osf.io/m7wnp) and Vicuna (osf.io/zfq5e) with 32 prime-target pairs consisting of a prime preamble followed by a target preamble. We measured whether ChatGPT completed a target preamble using a PO or DO structure (e.g., *The patient showed his hand to the nurse* vs. *The patient showed the nurse his hand*).

4a. DO-inducing prime preamble, different verb: *The racing driver gave the helpful mechanic ...*
4b. PO-inducing prime preamble, different verb: *The racing driver gave the torn overall ...*
4c. DO-inducing prime preamble, same verb: *The racing driver showed the helpful mechanic ...*
4d. PO-inducing prime preamble, same verb: *The racing driver showed the torn overall ...*
4e. Target preamble: *The patient showed ...*

We observed structural priming in both LLMs, with a higher proportion of PO completions of a target preamble when the corresponding prime preamble had been completed as a PO than when it had been completed as a DO (ChatGPT: 0.71 vs. 0.58, $β = 1.03$, $SE = 0.12$, $z = 8.68$, $p < .001$; Vicuna: 0.81 vs. 0.51, $β = 2.93$, $SE = 0.34$, $z = 8.70$, $p < .001$; see Fig 2 top left). Verb type (different vs. same verbs across prime and target) did not have an effect on completions for either model (ChatGPT: 0.66 vs. 0.63, $\beta$ = -0.06, $SE$ = 0.09, $z$ = -0.67, $p$ = .504; Vicuna: 0.66 vs. 0.66, $\beta$ = -0.14, $SE$ = 0.23, $z$ = -0.61, $p$ = .545). But, importantly, verb type interacted with prime structure, indicating a lexical boost, with a stronger priming effect when the prime and the target had the same verb (ChatGPT: $\beta$ = 0.40, $SE$ = 0.15, $z$ = 2.73, $p$ = .006: Vicuna: $\beta$ = 1.20, $SE$ = 0.45, $z$ = 2.68, $p$ = .007). These findings suggest that ChatGPT and Vicuna resemble humans in being susceptible to structural priming and the lexical boost.

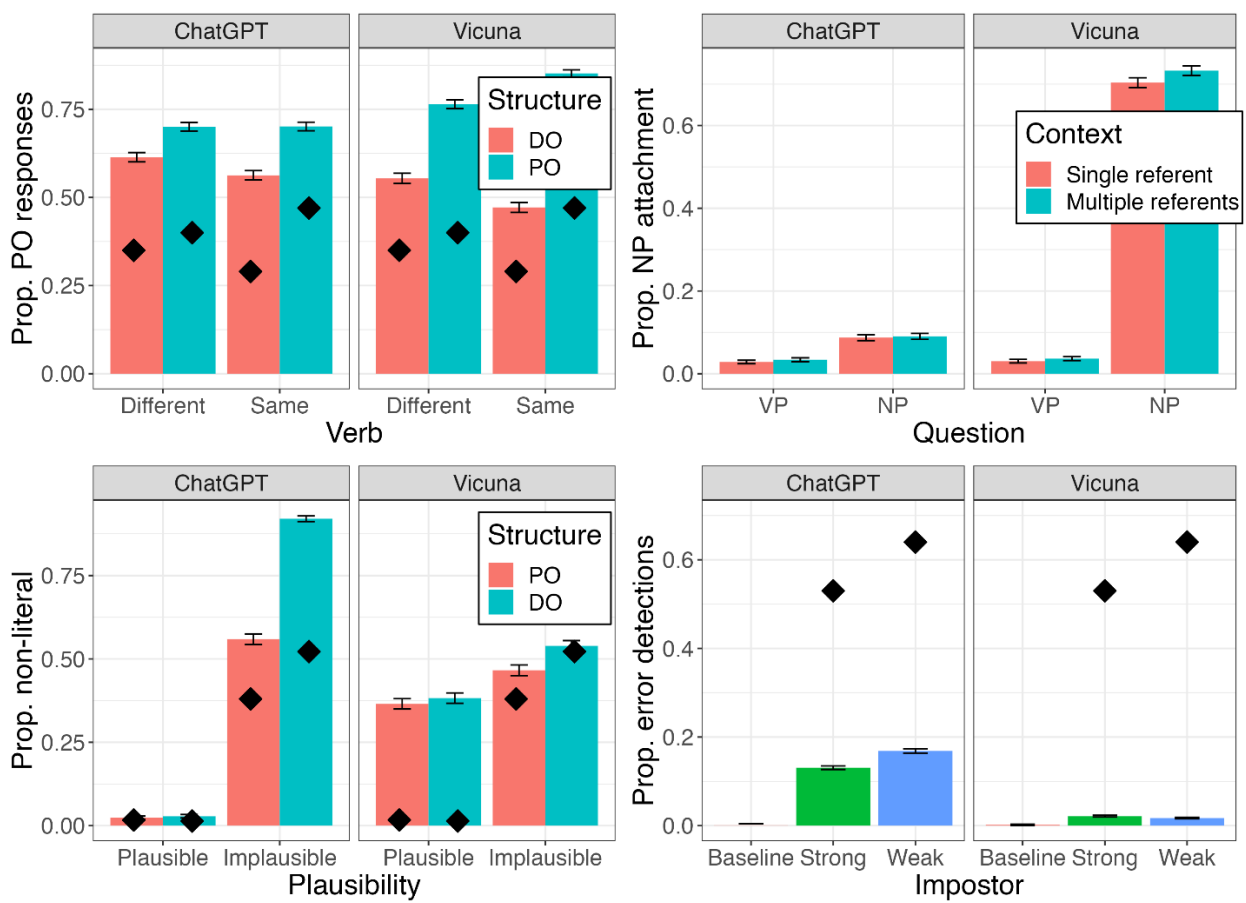

**Fig 2**. Results of structural priming (top left; human conditional means, shown as diamonds, are from Pickering & Branigan, 1998), syntactic ambiguity resolution (top right; no human conditional means available), implausible sentence interpretation (bottom left; human conditional means are from Gibson et al., 2013), and semantic illusions (bottom right; human conditional means are from Hannon & Daneman, 2001). Error bars stand for 95% confidence intervals. Note that in some cases, materials have been modified, so the human means are for comparison only.

*Syntax: syntactic ambiguity resolution*

In what is known as the verb phrase/noun phrase (VP/NP) ambiguity (e.g., *The ranger killed the dangerous poacher with the rifle*; see also 5a-d), people tend to interpret the syntactically ambiguous prepositional phrase (PP, *with the rifle*) as modifying the verb phrase (*kill the dangerous poacher*; VP attachment) rather than the noun phrase (*the dangerous poacher*; NP attachment) (e.g., Rayner et al., 1983; van Gompel et al., 2001). Critically, humans use contextual information to resolve the ambiguity and were more likely to have NP attachments when the discourse has introduced multiple possible referents for the NP (e.g., 5c/d) than a single referent (e.g., 5a/b) (Altmann & Steedman, 1988). We tested whether LLMs also use context to disambiguate the VP/NP ambiguity. After reading a discourse sentence (introducing a single referent or multiple possible referents for the critical NP) followed by a sentence containing the VP/NP ambiguity, ChatGPT/Vicuna answered a question regarding the ambiguous sentence (e.g., 5a-d). We also manipulated whether the question probes the VP attachment (e.g., 5a/c) or the NP attachment (e.g., 5b/d). If ChatGPT/Vicuna computes and retains a single semantic representation, we should find no difference in attachment (i.e., interpreting the PP as modifying the VP or the NP) between the VP and the NP probes; that is, if ChatGPT/Vicuna computes "hunter killing poacher using rifle", then it should answer "yes" to a VP probe and "no" to an NP probe (i.e., VP attachment). In this experiment, we presented ChatGPT (9 Jan 2023 version; osf.io/c28ur) and Vicuna (osf.io/2zpy7) 32 sets of sentences (e.g., 5a-d). We coded "yes" responses to VP and NP probes as VP attachment and NP attachment, respectively, and "no" responses as NP and VP attachment, respectively. Note that while the stimuli for the ChatGPT experiment always had an adjective modifying the critical NP (e.g., *the dangerous poacher*), the adjective was removed in the Vicuna experiment. This was done because the adjective serves to single out a referent for the NP when the discourse introduces multiple potential referents for the NP; thus, there would be no need to further use the PP to modify the NP, potentially leading to a lack of contextual effect.

> 5a. Single referent, VP probe: *There was a hunter and a poacher. The hunter killed the dangerous poacher with a rifle not long after sunset. Did the hunter use a rifle?*
> 5b. Single referent, NP probe: *There was a hunter and a poacher. The hunter killed the dangerous poacher with a rifle not long after sunset. Did the dangerous poacher have a rifle?*
> 5c. Multiple referents, VP probe: *There was a hunter and two poachers. The hunter killed the dangerous poacher with a rifle not long after sunset. Did the hunter use a rifle?*
> 5d. Multiple referents, NP probe: *There was a hunter and two poachers. The hunter killed the dangerous poacher with a rifle not long after sunset. Did the dangerous poacher have a rifle?*

Both models attached the ambiguous PP more often to the VP than to the NP (ChatGPT: 0.94 vs. 0.06, $\beta = -9.43$, $SE = 0.72$, $z = -13.04$, $p < .001$; Vicuna: 0.63 vs. 0.37, $\beta = -1.37$, $SE = 0.16$, $z = -8.35$, $p < .001$; see Fig 2 top right). There were similar NP attachments in the multiple-referent context and in the single-referent context (ChatGPT: 0.06 vs. 0.06, $\beta = -0.08$, $SE = 0.43$, $z = -0.18$, $p = .861$; Vicuna: 0.37 vs. 0.36, $\beta = 0.18$, $SE = 0.10$, $z = 1.87$, $p = .061$; note that the effect of context was significant in Vicuna when we removed exemplar items), but more NP attachments when answering an NP probe than when answering a VP probe (ChatGPT: 0.09 vs. 0.03, $\beta = 3.27$, $SE = 0.97$, $z = 3.36$, $p < .001$; Vicuna: 0.72 vs. 0.03, $\beta = 5.63$, $SE = 0.48$, $z = 11.78$, $p < .001$). There was no significant interaction between context and question (ChatGPT: $\beta = 0.13$, $SE = 0.74$, $z = 0.18$, $p = .861$; Vicuna: $\beta = -0.16$, $SE = 0.24$, $z = -0.66$, $p = .511$). These findings suggest, first of all, that neither ChatGPT nor Vicuna used contextual information to resolve syntactic ambiguities (at least the VP/NP ambiguity) as humans do and they might retain multiple representations of the ambiguous sentence (i.e., treating *with the rifle* as potentially modifying both *the poacher* and *kill the poacher*).

*Meaning: implausible sentence interpretation*

Listeners sometimes have to recover an intended message from noise-corrupted input (Gibson et al., 2013; Levy et al., 2009). For example, an error in production or comprehension may turn a plausible sentence into an implausible one when a word is omitted (e.g., *to* being omitted from a plausible PO such as 6b, resulting in an implausible DO such as 6c) or when a word gets inserted (e.g., *to* being inserted into a plausible DO such as 6a, resulting in an implausible PO such as 6d). If people believe that an implausible sentence results from a plausible sentence being noise-corrupted, then they can interpret the implausible sentence nonliterally to recover the intended message (e.g., interpreting *the daughter* as the recipient in 6c/d). Gibson et al. (2013) showed that people nonliterally interpret implausible DO sentences more often than implausible PO sentences, probably because they believe that omissions of *to* are more likely than insertions of *to*. We presented ChatGPT (15 Dec 2022 version; osf.io/2pktf) and Vicuna (osf.io/a76f4) with 20 sentences (plausible or implausible, in a DO or PO structure), each followed by a yes/no question (e.g., 6e) probing whether the sentence is literally or nonliterally interpreted.

6a. Plausible DO: *The mother gave the daughter the candle.*
6b. Plausible PO: *The mother gave the candle to the daughter.*
6c. Implausible DO: *The mother gave the candle the daughter.*
6d. Implausible PO: *The mother gave the daughter to the candle.*
6e. Question: *Did the daughter receive something/someone?*

ChatGPT made more nonliteral interpretations for implausible than plausible sentences (0.74 vs. 0.03, $\beta = 10.85$, $SE = 0.73$, $z = 14.80$, $p < .001$; see Fig 2 bottom left), whereas the difference did not reach significance for Vicuna (0.50 vs. 0.37, $\beta = 2.20$, $SE = 1.24$, $z = 1.77$, $p = .076$). There was an effect of structure on interpretation in ChatGPT, with more nonliteral interpretations for DO than PO sentences (0.47 vs. 0.29, $\beta = 1.15$, $SE = 0.58$, $z = 1.94$, $p = .047$), but not in Vicuna (0.46 vs. 0.42, $\beta = 0.04$, $SE = 0.36$, $z = 0.11$, $p = .910$). The interaction between plausibility and structure was significant such that the increase in nonliteral interpretations for the DO structure compared to the PO structure was larger when a sentence was implausible than when it was plausible in both ChatGPT ($\beta = 4.47$, $SE = 1.17$, $z = 3.81$, $p < .001$) and in Vicuna ($\beta = 1.40$, $SE = 0.69$, $z = 2.02$, $p = .043$; this interaction failed to reach significance when we removed exemplar items). Critically, when we examined the implausible sentences alone, there was humanlike pattern of interpretations in ChatGPT, with more nonliteral interpretations for implausible DO sentences than for implausible PO sentences (0.92 vs. 0.56, $\beta = 3.40$, $SE = 0.74$, $z = 4.59$, $p < .001$) but not in Vicuna (0.54 vs. 0.47, $\beta = 0.77$, $SE = 0.57$, $z = 1.35$, $p = .178$). These findings suggest that ChatGPT (but not Vicuna) was sensitive to syntactic structure, like humans, in the interpretation of implausible sentences.

*Meaning: semantic illusions*

People often fail to notice what seem to be conspicuous errors in sentences. For example, when asked the question in 7b, many people do not notice that Snoopy, from the comic strip *Peanuts*, is a not a cat but a dog. People are more likely to notice an erroneous word when it is semantically less similar to *dog*, such as *mouse* in 7c (Erickson & Mattson, 1981). Such semantic illusions suggest that representing word meanings while processing sentences involves partial matches in semantic memory (Reder & Kusbit, 1991). We asked ChatGPT (9 Jan 2022 version;

osf.io/r67f2) and Vicuna (osf.io/s3uwy) trivia questions that contained a semantically appropriate keyword, a strong (semantically closely related) impostor, or a weak impostor, with a total of 54 sentences in three conditions, taken from Hannon & Daneman (2001). Following Erickson and Mattson (1981) and Hannon and Daneman (2001), we instructed the models either to answer the question or, if they detected a semantic error (which we illustrated with an example), to say *wrong* (i.e., to give an error report).

> 7a. Baseline: *Snoopy is a black and white dog in what famous Charles Schulz comic strip?*
> 7b. Strong imposter: *Snoopy is a black and white cat in what famous Charles Schulz comic strip?*
> 7c. Weak imposter: *Snoopy is a black and white mouse in what famous Charles Schulz comic strip?*

For ChatGPT, compared to the baseline condition, there were more error reports in the strong impostor condition (0.00 vs. 0,13, $\beta = 0.87$, $SE = 0.00$, $z = 122035$, $p < .001$; see Fig 2 bottom right) and in the weak impostor condition (0.00 vs. 0.17, $\beta = 2.83$, $SE = 0.00$, $z = 677303$, $p < .001$); critically, more errors were reported in the weak than strong imposter condition (0.17 vs. 0.13, $\beta = 1.71$, $SE = 0.82$, $z = 2.10$, $p = .036$; though this difference did not reach significance when removing exemplar items). For Vicuna, there statistically similar proportions of error reports between the baseline and the strong imposter condition (0.002 vs. 0.022, $\beta = -3.01$, $SE = 1.65$, $z = -1.82$, $p = .069$) and between the baseline and the weak imposter condition (0.002 vs. 0.017, $\beta = 0.82$, $SE = 1.27$, $z = 0.65$, $p = .517$); interestingly, we observed significantly more error reports in the weak than strong imposter condition ($\beta = 3.96$, $SE = 1.31$, $z = 3.02$, $p = .003$), though numerically the mean error report rate was lower in the weak than strong imposter condition (0.017 vs. 0.022). These findings that ChatGPT, but not Vicuna, has the humanlike tendency to gloss over a conspicuous error caused by an expression that is semantically similar to the intended expression.

*Discourse: implicit causality*
The verb determines the thematic roles of the subject and the object in a sentence (e.g., the subject *Gary* and the object *Anna* serve respectively as the *stimulus* and the *experiencer* in 8a but respectively as the *experiencer* and the *stimulus* in 8b). Some verbs lead people to attribute causality to either the subject or the object (Brown & Fish, 1983; Garvey & Caramazza, 1974). For example, a stimulus-experiencer verb such as *scare* often leads people to attribute causality to the subject (e.g., completing the preamble in 8a with *he was violent*) while an experiencer-stimulus verb such as *fear* often leads people to attribute causality to the object (e.g., completing the preamble in 8b with *she was violent*). We asked ChatGPT (30 Jan 2023 version and 13 Feb 2023 version; osf.io/jx2sy) and Vicuna (osf.io/ac3qd) to complete sentences adapted from Fukumura & van Gompel (2010), manipulated to elicit pronouns referring to either subject or objects, with 32 sentences in two conditions.

> 8a. Stimulus-experiencer verb: *Gary scared Anna because...*
> 8b. Experiencer-stimulus verb: *Gary feared Anna because...*

Both LLMs more often completed a sentence with a pronoun referring to the object (e.g., *Gary scared/feared Anna because she/he was violent*) following an experiencer-stimulus verb such as *fear* than following a stimulus-experiencer verb such as *scare* (ChatGPT: 0.95 vs. 0.00, $\beta$ = 14.17, $SE$ = 0.94, $z$ = 15.11, $p$ < .001; Vicuna: .89 vs. 0.01, $\beta$ = 14.95, $SE$ = 1.57, $z$ = 9.51, $p$ < .001; see Fig 3 top left). These findings suggest that ChatGPT and Vicuna are sensitive to the semantic biases of verbs, as humans are.

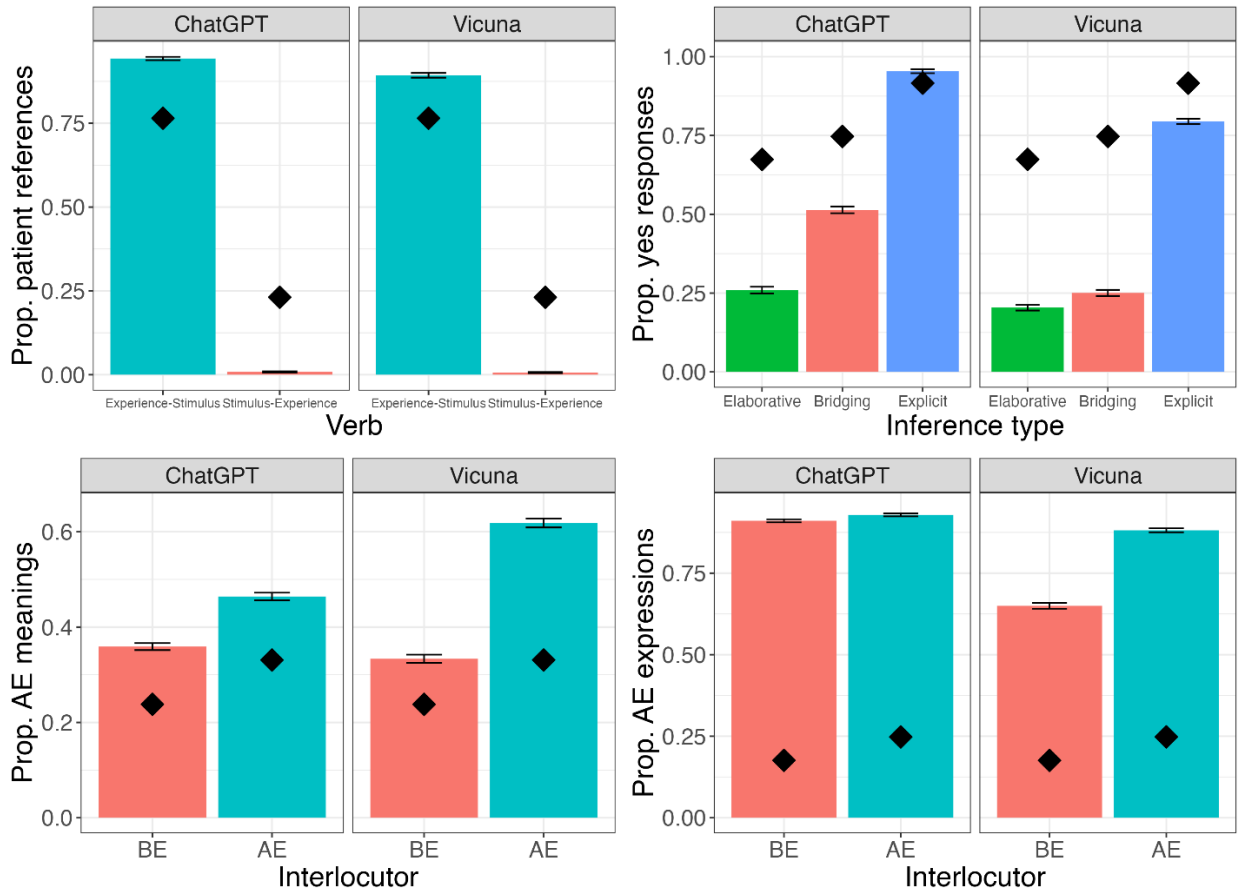

**Fig 3**. Results of implicit causality (top left; human conditional means, shown as diamonds, are from Fukumura & van Gompel, 2010), drawing inferences (top right; human conditional means are from Singer & Spear, 2015), interlocutor-sensitive word meaning access (bottom left; human conditional means are from Cai et al., 2017), and interlocutor-sensitive lexical retrieval (bottom right; human conditional means are from Cai et al., accepted in principle). Error bars stand for 95% confidence intervals. Note that in some cases, materials have been modified, so the human means are for comparison only.

*Discourse: drawing inferences*

People make bridging inferences, which connect two pieces of information, more often than they make elaborative inferences, which extrapolate from a single piece of information (Singer & Spear, 2015). For example, people are likely to infer that Sharon cut her foot after reading 9b but less likely to do so after reading 9c. We presented ChatGPT (9 Jan 2023 version; osf.io/e3wxc) and Vicuna (osf.io/zs9wb) with a short passage and a yes/no question, with 24 items based on the design of Singer and Spear (2015) and using materials adapted from McKoon and Ratcliff (1986). A passage either contained explicit information, required a bridging inference, or required an elaborative inference. As all 24 target items were likely to elicit "yes" responses, we also presented the models with 24 fillers designed to elicit "no" responses.

> 9a. Explicit: *While swimming in the shallow water near the rocks, Sharon cut her foot on a piece of glass. She had been looking for the watch that she misplaced while sitting on the rocks.*
> 9b. Bridging: *While swimming in the shallow water near the rocks, Sharon stepped on a piece of glass. She called desperately for help, but there was no one around to hear her.*
> 9c. Elaborative: *While swimming in the shallow water near the rocks, Sharon stepped on a piece of glass. She had been looking for the watch that she misplaced while sitting on the rocks.*
> Question: *Did she cut her foot?*

Both LLMs produced fewer "yes" responses in the bridging condition than in the explicit condition (ChatGPT: 0.51 vs. 0.95, $\beta = -5.06$, $SE = 0.10$, $z = -50.16$, $p < .001$; Vicuna: 0.25 vs. 0.79, $\beta = -4.32$, $SE = 0.50$, $z = -8.65$, $p < .001$; see Fig 3 top right) and fewer "yes" responses in the elaborative than explicit condition (ChatGPT: 0.26 vs. 0.95, $\beta = -7.40$, $SE = 0.12$, $z = -62.68$, $p < .001$; Vicuna: 0.20 vs. 0.79, $\beta = -4.41$, $SE = 0.41$, $z = -10.73$, $p < .001$). Critically, ChatGPT gave fewer "yes" responses in the elaborative than bridging condition (0.26 vs. 0.51, $\beta = -2.87$, $SE = 0.58$, $z = -4.93$, $p < .001$), whereas Vicuna gave similar "yes" responses for the bridging and elaborative conditions (0.25 vs. 0.20, $\beta = -0.09$, $SE = 0.42$, $z = -0.22$, $p = .830$). These findings suggest that ChatGPT, but not Vicuna, is less likely to make elaborative than bridging inferences, as humans are.

*Interlocutor sensitivity: word meaning access*

Words and other expressions may mean different things to different people. For example, speakers of British English (BE) typically interpret *bonnet* as referring to a car part, while speakers of American English (AE) typically interpret *bonnet* as referring to a hat, and listeners take such demographic attributes of speakers into account when comprehending language (e.g., Cai et al., 2017; Van Berkum et al., 2008). For instance, Cai et al. (2017) showed that BE-speaking participants were more likely to access AE meanings of cross-dialectally ambiguous words (e.g., *bonnet, gas*) when the words were spoken in an AE than a BE accent.

ChatGPT and Vicuna, at the time of testing, did not take spoken input, so we manipulated the interlocutor's dialectal background by explicitly telling ChatGPT (15 Dec 2022 version; osf.io/k2jgd) and Vicuna (osf.io/zs9wb) that the interlocutor was a BE speaker (*Hi, I am a British English speaker. I am from the UK. I am now living in London and studying for a BA degree at King's College London*) or an AE speaker (*Hi, I am an American English speaker. I am from the USA. I am now living in New York and studying for a BA degree at the City University of New York*). We then presented, one at a time, 36 cross-dialectally ambiguous words (taken from Cai et al., 2017) and asked ChatGPT and Vicuna to give an associate to each word.

We coded whether the models accessed the BE or AE meaning of these words based on the associates it gave (e.g., "hat" as an associate to *bonnet* would suggest that ChatGPT accessed the word's AE meaning). There was more access to the AE meaning of a target word when the interlocutor was introduced as an AE speaker than a BE speaker, in both ChatGPT (0.46 vs. 0.36, $β = 1.85$, $SE = 0.26$, $z = 7.14$, $p < .001$; see Fig 3 bottom left) and Vicuna (0.62 vs. 0.33, $β = 2.80$, $SE = 0.54$, $z = 5.15$, $p < .001$). These findings suggest that both models are sensitive to the user's dialectic background in understanding word meanings.

*Interlocutor sensitivity: lexical retrieval*

People can take a listener's dialectal background into account when retrieving words during language production (Cai et al., accepted in principle; Cowan et al., 2019). Using a word puzzle game, Cai et al. (accepted in principle) gave participants a definition spoken in either a BE or AE accent and asked them to type the defined word/phrase. Critically, the expected words differed between BE and AE for some of the definitions (e.g., *a housing unit common in big cities that occupies part of a single level in a building block* defines the word *flat* in BE and the word *apartment* in AE). Cai et al. found that participants produced more AE expressions for definitions spoken by an AE speaker than by a BE speaker.

In the experiment, we told ChatGPT (15 Dec 2022 version; osf.io/28vt4) and Vicuna (osf.io/ryjzv) that the interlocutor was a BE or AE speaker (using the same introductions as in the word meaning access experiment). The interlocutor gave a definition of a word/phrase and the LLM supplied the defined word/phrase. There were more AE expressions supplied when the LLM was told that the definitions came from an AE speaker than from a BE speaker, for both ChatGPT (0.93 vs. 0.91, $β = 4.39$, $SE = 1.56$, $z = 2.81$, $p = .005$; see Fig 2 bottom right) and Vicuna (0.88 vs. 0.65, $β = 3.54$, $SE = 0.51$, $z = 6.91$, $p < .001$). These findings suggest that both models are sensitive to the user's dialectic background in their lexical choices.

**DISCUSSION**

LLMs and LLM-based chatbots such as ChatGPT and Vicuna respond to questions and continue prompts by breaking text down into tokens and, based on the hundreds of billions of tokens they have been trained on, predict which token will come next (Brown et al., 2020; Ouyang et al., 2022). Despite the impressive engineering success of LLMs, it remains a mystery whether LLMs resemble humans in language use. Our experiments investigated whether LLM language use displays humanlike regularities in language comprehension and production. They showed that ChatGPT replicated human patterns in language comprehension and production in 10 out of 12 psycholinguistic tasks and Vicuna in 7 out of the same 12 tasks. These findings suggest that both models largely approximate human language processing. In particular, both LLMs associated certain word forms with certain semantic features (i.e., sound-shape and sound-gender associations), updated their lexical-semantic and syntactic representations based on recent input (i.e., word-meaning and structural priming), continued discourse coherently according to the semantics of verbs (i.e., implicit causality), and was sensitive to the identity of its interlocutor in word-meaning access and word production (i.e., interlocutor-sensitive word-meaning access and lexical retrieval). ChatGPT, but not Vicuna, was sensitive to the likelihood of errors when computing implausible sentences, displayed humanlike semantic illusion, and drew inferences from sentences (i.e., inference drawing). Neither ChatGPT nor Vicuna used contextual information to resolve syntactic ambiguities (i.e., contextual modulation of syntactic ambiguity resolution), or chose the shorter form of a word when conveying more predictable information (i.e., predictivity effects on word length). Though in many experiments, the strength of the effect appears to differ substantially from what has been observed in humans (see bars vs. diamonds in Figures 1, 2 and 3), it should be noted that, in some experiments, we modified the stimuli and/or have access only to condition-level means, which preclude direct comparisons (but see

Supplemental Information for some direct comparisons between language models and humans). It is improbable that the LLMs exhibited humanlike behaviors solely by "learning" them from research papers that reported these effects, as we obtained (almost) identical patterns of results when excluding the specific stimuli mentioned in a relevant research paper (refer to Supplemental Information for more details). In what follows, we first discuss how these human behaviours may be achieved in LLMs and then consider potential reasons the LLMs diverged from humans in some cases.

Some replications of human linguistic behaviours may be explained by the computational implementations of LLMs. Both ChatGPT and Vicuna are built on transformer architectures (Vaswani et al., 2017), which allow them to vary how much weight they assign to different tokens within recent conversation history when predicting the subsequent token. This context sensitivity means that LLMs represent the same word differently in different contexts (Ethayarajh, 2019; Peters, 2018), so recent usage of an ambiguous word (e.g., *post*) influences its representation, which in turn influences the likelihood of producing associates related to one meaning or the other (e.g., *job* versus *mail*), thereby approximating human behaviour (Rodd et al., 2013). Similarly, recent exposure to *American English* or *New York*, for instance, could influence the set of tokens likely to follow a cross-dialectically ambiguous word (e.g., *bonnet*) and so increase the likelihood that an LLM will supply associates related to the American rather than British meaning of a cross-dialectically ambiguous word, again approximating human behaviour (Cai et al., 2017). Context also influenced ChatGPT's responses across adjacent sentences. It tended to infer, for example, that *step on glass* plus *call out for help* more often implies "foot cutting" than *step on glass* alone, which suggests that it accumulates evidence, as humans do (Singer & Spear, 2015). Within sentences, the LLMs might weight objects more heavily than subjects following experiencer-stimulus verbs, such as *feared*, and vice versa following stimulus-experiencer verbs, such as *scared*. This could explain how they come to refer to objects versus subjects depending on verb type in a humanlike fashion (e.g., Brown & Fish, 1983).

LLMs rely on sub-word tokens of varying lengths to represent low-frequency words, which helps them process words that do not occur in their training data (Radford et al., 2019). For example, GPT-3.5 and -4 divide the non-word *kiki* and the made-up name *Pelcrad* into two and three tokens, respectively (*k-iki*; *Pel-cr-ad*). If tokens such as *k* co-occur with references to spikiness more often than to roundness, then that may explain how ChatGPT produces humanlike sound symbolic associations. Similarly, if tokens such as *ad* occur at the end of masculine English names more often than feminine English names, that may explain why ChatGPT/Vicuna more often uses masculine pronouns to refer to made-up names that contain such tokens. Alternatively, sound symbolic associations might be an emergent quality of LLMs, rather than being reducible to tokens, given that LLMs can identify what letters their tokens contain at above-chance levels (Kaushal & Mahowald, 2022). Both explanations could explain analogies between LLMs and findings that patterns in word forms help people represent the meanings of unfamiliar words (e.g., Cassani et al., 2020; Gatti et al., 2023; Haslett & Cai, 2023).

The fact that LLMs change semantic representations of words to fit contexts (Ethayarajh, 2019) may help to account for ChatGPT's humanlike susceptibility to semantic illusions (e.g., answering *Peanuts* in response to *Snoopy is the black-and-white cat in what Charles Schulz comic strip?*). Perhaps ChatGPT and Vicuna detect far fewer errors than humans do (see Fig. 2) because congruence with context overrides the context-independent representations that would otherwise identify words as anomalous. Similarly, context-sensitive semantic representations may lead LLMs to interpret sentences nonliterally (Brown et al., 2020), such as treating *the daughter* as the recipient in the implausible sentence *The mother gave the candle the daughter*. What is intriguing is that ChatGPT (though not Vicuna) nonliterally interprets implausible DO sentences more often than PO ones, as humans do, though this effect is far greater in ChatGPT than in humans (see Fig. 2 and Supplemental Information). Humans are argued to estimate the likelihood of a sentence being implausible due to noise corruption (e.g., words being accidentally omitted or inserted; Cai et al., 2022; Gibson et al., 2013). It is more likely for people to accidentally omit a word in a plausible PO sentence (e.g., in *The mother gave the candle to the daughter*, the likelihood of a word such as to being accidentally omitted can be assumed to be about 1/8) to result in an implausible DO than for people to insert a word into a plausible DO (e.g., the likelihood is extremely low for people to accidentally insert *to*, rather than any other words, into sentence *The mother gave daughter the candle*) to result in an implausible PO; it is thus suggested that humans estimate the likelihood of a sentence being implausible due to noise corruption (Cai et al., 2022; Gibson et al., 2013). ChatGPT seems to have a humanlike estimation regarding the likelihood of a sentence being implausible due to noise corruption, though how it arrives at this estimation requires further investigation.

There has been evidence that LLMs have the tendency to re-use the syntactic structure of a recently produced sentence (structural priming; see also Michaelov et al., 2023; Prasad et al., 2019; Sinclair et al., 2022). Note that LLMs produce sentences by predicting upcoming tokens one at a time, evaluating the probabilities of potential candidates as a function of previously produced tokens but, crucially, not yet-to-be-produced tokens. For instance, when determining the next word to continue a preamble such as *The patient showed…*, an LLM may evaluate the probabilities of theme-like nouns (e.g., *hand*) as the next word (thus resulting in a PO structure in the eventual sentence) or recipient-like nouns (e.g., *nurse*) as the next word (thus resulting in a DO structure); however, without the assumption of some syntax-like representations, there is no way for a model to prefer a theme-like or recipient-like noun as the next word in its continuation. A more likely scenario is that LLMs learn supra-token representations, which could be syntactic (e.g., DO or PO) or semantic (e.g., a verb-theme-recipient vs. verb-recipient semantic frame) (see also Cai & Zhao, 2024). Thus, as in humans, when a model has previously completed a preamble into a particular sentence (e.g., DO or PO), the model has the supra-token syntax-like representation in memory, leading it to be more likely to be used in subsequent continuations. Structural priming is further enhanced when verbs are repeated across the prime and the target; it is possible that verbs (or lexical heads in general) are linked to supra-token syntax-like (e.g., Pickering & Branigan, 1998) or serve as additional cues (e.g., van Gompel et al., 2023), thus leading to enhanced structural priming.

We observed that ChatGPT replicated more humanlike patterns of language use than Vicuna did (10 versus 7 out of the 12 experiments). Given that increasing model size improves performance (e.g., Devlin et al., 2019), we assume that this difference in mimicking the nuances of human language use should be attributed to Vicuna being a smaller model than GPT-3.5 (the presumed LLM underlying ChatGPT for each of our experiments, conducted between the December 2022 and February 2023). The Vicuna model we tested comprises 13 billion parameters and was trained on 370 million tokens of text, whereas GPT-3.5 is speculated to comprise 175 billion parameters and to have been trained on 499 billion words of text (e.g., Hughes, 2023), though OpenAI has not revealed the actual model size, and estimations vary (e.g., Farseev, 2023). Additionally, GPT-3.5 was fine-tuned using human feedback, an approach that likely led to its more refined and contextually accurate responses (Christiano et al., 2017;

OpenAI, 2022). This fine-tuning involves adjusting the model based on human input, facilitating more accurate and context-appropriate outputs. In contrast, Vicuna was fine-tuned using over 70,000 human-ChatGPT dialog exchanges, which specialized the model for conversational contexts. Thus, ChatGPT's broader training and fine-tuning processes with human feedback give it an advantage in producing more nuanced and contextually relevant responses (see also Sun et al., 2023).

In two of the experiments, neither ChatGPT nor Vicuna replicated the patterns of human participants. First, they did not select shorter forms of words (e.g., *math* versus *mathematics*) when those meanings were predictable (unlike Mahowald et al., 2013). There is evidence from large-scale text analyses that words which are more predictable tend to be shorter (Piantadosi et al., 2011), so given that LLMs are trained to predict upcoming tokens, we would expect them to be sensitive to this trend. One possible explanation for the null effect is that tokenization interferes with patterns involving word length, since longer words are more often segmented into more tokens. Alternatively or additionally, LLMs sometimes segment the short and the long forms into very different tokens (e.g., GPT-4 segments *roach* into "ro" and "ach" and *cockroach* into "cock" and "roach"). As a result, while humans may treat the short and long words as variants of nearly identical meanings and differentially use the variants according to their informativeness, LLMs may not. Of course, it is also possible the null effect instead stems from ChatGPT's aversion to short forms, which it produces far less often than humans do (see Fig. 1).

Second, both models failed to take context into account when resolving the VP/NP syntactic ambiguity (e.g., *The hunter killed the dangerous poacher with a rifle*). Human participants are more likely to treat the ambiguous phrase *with a rifle* as modifying *the dangerous poacher* if the discourse has introduced multiple poachers than if it has only introduced one poacher (Altmann & Steedman, 1988), a pattern we did not observe with either LLM (though note that our contextual manipulation was different from that in the Altman & Steedman study). Again, this is surprising given the LLMs' superb ability to make use of contextual information in language comprehension and production. One potential explanation is that, in our stimuli for ChatGPT, the critical noun was already modified (e.g., *the dangerous poacher*), thus singling out the intended referent and making it less necessary for ChatGPT to treat the *with*-phrase as a further modification of the critical noun. However, removing the adjective in the Vicuna experiment did not result in a contextual modulation. The lack of sensitivity to contextual information in syntactic parsing is reminiscent of a similar absence of contextual effects in pragmatic understanding observed in ChatGPT (Qiu et al., 2023). It is also worth pointing out that ChatGPT and Vicuna may make multiple interpretations corresponding to the alternative syntactic analyses (e.g., both the hunter and the poach may have the rifle), thus tending to answer "yes" to both an NP probe (e.g., *Did the hunter have the rifle?*) and a VP probe (e.g., *Did the dangerous poacher have a rifle?*). Of course, it might also be possible that these models simply have a bias to answering "yes" to a question (e.g., Dentella et al., 2023).

In conclusion, LLMs have demonstrated proficiency across a wide range of linguistic tasks, such as summarization, text classification, translation, and question-answering (e.g., Radford et al., 2018, 2019; Wei et al., 2022). Our study further showed that ChatGPT, and Vicuna to a lesser extent, resemble humans in performance on many psycholinguistic tasks. Assuming that ChatGPT and Vicuna were unlikely to have received explicit instructions or human feedback on these patterns of language use, one can reasonably conclude that these abilities emerge when a transformer with billions of parameters is trained on hundreds of billions of words. If so, this raises the interesting possibility that LLMs such as ChatGPT (and Vicuna to a lesser extent) can be used, by psycholinguists and cognitive psychologists, as models of language users (e.g., Aher et al., 2023; Argyle et al., 2023; Jain et al., 2023). Perhaps researchers can experiment with LLMs to generate hypotheses (which, if confirmed in an LLM, can be later tested on human subjects), assess the replicability of existing psycholinguistic effects (and using discrepancies to motivated replication attempts with humans), estimate effect sizes (which can inform statistical power estimation in human experiments), and model language development (which could aid teachers and speech-language pathologists). At the same time, it is essential to proceed with caution. Humanlike performances on psycholinguistic tasks suggest that LLMs approximate human language processing, but responses to prompts do not entail thought or intent (Bender & Koller, 2020; Mahowald et al., 2023). LLMs are prone to hallucinations (e.g., Ji et al., 2023; Xu et al., 2024) and as some of our experiments demonstrated, they regularly fail to notice mistakes by others. Humans therefore have an obligation to scrutinize this tool and to take accountability for how they use it.

**MATERIALS AND METHODS**

All experiments were preregistered (ChatGPT: osf.io/vu2h3/registrations; Vicuna: osf.io/sygku/registrations), with all materials and analytical plans preregistered prior to data collection and analysis. We ran the ChatGPT experiments with a web interface (https://chat.openai.com/) and the Vicuna experiments with the model's API. For ChatGPT, we adopted a multiple-trial-per-run design, as with a human participant (i.e., there were multiple trials in each session/run with ChatGPT); such a design was adopted because it reduced the number of runs/sessions as at the time of testing it was sometimes difficult to secure a session with ChatGPT. With Vicuna, we used a one-trial-per-run design, where we only presented the experimental instructions and one target trial in each run/session with the model. Thus, in the Vicuna experiments there was no need for filler trials and the fillers in the following description were only for the ChatGPT experiments.

Unless otherwise stated, all experiments shared some common procedures, as specified in the preregistrations. First, all ChatGPT experimental materials were assigned to different lists according to the number of within-item conditions (e.g., two lists if there were two within-item conditions) such that different experimental versions of the same item appeared in different lists; all stimuli (targets and fillers) in a list were randomly presented; note that in Vicuna experiments there was only one trial per run so no lists were needed). Second, we used a Python script to simulate a human interlocutor having a chat with ChatGPT/Vicuna. The simulated interlocutor always began with instructions regarding how the task was to be done. Third, each item in an experiment was run 1000 times with ChatGPT/Vicuna (in ChatGPT, the stimuli in a list); in our pilot, we found that ChatGPT tended to stop responding after a certain number of prompts, so for experiments with more than 70 trials, we split the stimuli into two blocks and ran each block 1000 times. If an experimental run ended prematurely, the run was replaced.

*Sounds: sound-shape association*

The experiment began with the following instructions:

*Hi, I'd like to play a NON-WORD guessing game with you. You need to guess whether the non-word refers to a round or spiky shape, based on its pronunciation. If*

*you don't know the meaning, just guess the shape. Please don't ask any questions. For each non-word, please say only "round" or "spiky". Is that OK?*

There were 20 trials, 10 with a novel word deemed spiky-sounding by human participants (Sidhu & Pexman, 2017) and 10 with a round-sounding novel word (osf.io/6wxp3); in each trial, we presented a novel word (e.g., *tuhkeetee*) and ChatGPT/Vicuna decided whether it referred to a round or spiky shape. We used a Python script to automatically extract "round" and "spiky" from the responses. Responses where automatic text extraction failed to detect a "round" or "spiky" response or where it detected both a "round" and a "spiky" response were coded by a native English speaker (as "round" or "spiky", or, if neither or both apply, as "other") in a condition-blind manner. Sometimes ChatGPT provided a justification or elaboration for its answer (e.g., *Based on the pronunciation of the non-word "tuhkeetee," I would guess that it refers to a spiky shape. However, without more information, it is difficult to be certain. Please provide a different non-word or give me more information about the shape you are thinking of*); in this case, we used the shape judgement but ignore the elaboration. We excluded "other" responses from the analysis (0.5% and 2.6% of all the data respectively for ChatGPT and Vicuna). For the follow-up correlation analysis, we had human means for 10 round-sounding items but only 8 spiky-sounding items because Sidhu and Pexman (2017) did not use one spiky-sounding word (*puhkeetee*) in the corresponding experiment and because another item (*puhtay*) elicited "spiky" judgements from humans only 42% of the time, so we replaced it (with *keepa*).

*Sounds: sound-gender association*
The experiment began with the following instructions:

*I'd like to play a sentence completion game with you. I will provide a fragment and I would like you to repeat the fragment and complete it into a full sentence.*

There were 16 target trials and 16 filler trials (osf.io/7yrf8). In a target trial, we presented a preamble that contained a novel name as the subject of the preamble (e.g., *Although Pelcrad was sick …*) and ChatGPT/Vicuna completed the preamble into a full sentence (e.g., *Although Pelcrad was sick, he got up and went to work*). We determined whether ChatGPT/Vicuna referred to the novel name as feminine or masculine by first automatically extracting pronouns (*she/her/hers* or *he/him/his*) from ChatGPT/Vicuna completions. For responses where no pronoun or multiple pronouns of different genders were detected, we had a native speaker of English determine if the novel name was referred to as feminine or masculine. If a response was judged to refer to the novel name as neither feminine nor masculine, or not to refer to the novel name at all, then it was coded as an "other" response and was excluded from further analyses (24.3% and 7.8% of all the data respectively for ChatGPT and Vicuna).

*Words: word length and predictivity*
The experiment began with the following instructions:

*Hi, I'd like to play a sentence completion game with you. I will provide a sentence preamble and two choices of words to complete the preamble. Please choose a word that you think best completes the sentence. For instance, if you are given the*

*following preamble and choices: The boy went to the park to fly a ... 1. plane 2. kite. You can choose "kite" as a completion. Just give me the one word that you choose. Shall we start?*

The stimuli were the same as in Mahowald et al. (2013), consisting of 40 target items and 40 fillers (osf.io/n645c), divided into two blocks (10 targets and 10 fillers in each block). In a trial, we presented ChatGPT with a sentence preamble with the last word missing and ChatGPT/Vicuna chose between two words (e.g., *Susan was very bad at algebra, so she hated… 1. math 2. mathematics.*). For ChatGPT, the order of the two choices was counter-balanced across lists (i.e., the order of the long and short candidate words was counterbalanced: On each run, we presented ChatGPT with one of two lists, each containing one order for each item, and 20 short-first and 20 long-first stimuli). We coded whether ChatGPT/Vicuna chose the short or long word in a target trial.

*Words: word meaning priming*

The experiment consisted of two parts: a priming part and a word association part. The priming part had the following instructions:

*I would like to present you with a list of unrelated sentences. Please just read them; you don't have to do anything with them for now. Is that OK?*

We presented a set of 44 sentences in one go to ChatGPT/Vicuna, including 13 word-meaning primes, 13 synonym primes and 18 filler sentences (osf.io/ym7hg); note that when a target word was in the no-prime condition, there was no prime sentence in the priming part. This was immediately followed by the word association part (for ChatGPT, all the 39 ambiguous words were presented one by one in a random order on a run; for Vicuna, only one ambiguous word was presented on a run), with the following instructions:

*Next, I am going to present a list of unrelated words one by one; upon reading a word, please provide ONLY ONE word/phrase as an associate. For instance, if I say "milk", you can provide "breakfast" or "cow" as an associate. Is that OK?*

For each of 39 target ambiguous words (e.g., *post*), ChatGPT gave an associate (e.g., *mail*). We used the algorithm and database developed by Gilbert and Rodd (2022) to code whether an associate related to the (primed) subordinate meaning of a target word. There were 516 unique target-associate pairs not available in the database (50.2% of all unique pairs), two native speakers of English independently and condition-blindly coded whether an associate related to the subordinate meaning of the target word. Coding disagreements between the two coders (9.7% of manually-coded pairs) were resolved by a third coder, also a native speaker, in a condition-blind manner failed to provide an associate were coded as "other" and removed from further analyses (0.2% and 0% of all the data respectively for ChatGPT and Vicuna).

*Syntax: structural priming*

This experiment was run concurrently with the ***implicit causality*** experiment for ChatGPT (but not for Vicuna) because they had the same task and their target stimuli could serve as filler stimuli to each other. The experiment began with the following instructions:

> *I'd like to play a sentence completion game with you. I will provide a sentence preamble and I would like you to repeat the preamble and continue it into a full sentence.*

There were 64 preambles, forming 32 prime-target pairs, together with 64 filler preambles, 32 of which were experimental stimuli for the concurrent experiment (osf.io/k3cfv). For ChatGPT, these stimuli were divided into two blocks. In each pair, the prime (e.g., *The racing driver showed the helpful mechanic ...*) was always presented first for ChatGPT/Vicuna to complete (e.g., *The racing driver showed the helpful mechanic the problem with the car, hoping they would be able to fix it in time for the next race*), followed by the target preamble. For data coding, we made use of a pre-trained language model named "en_core_web_trf" (https://spacy.io/models/en) to generate dependency labels for the arguments of a verb. We specified all the verbs in model responses. The algorithm determined whether a response had a particular structure depending on the labels of the verb's arguments. To test the accuracy of the automatic coding using the algorithm, we did 5 pilot runs of the structural priming experimental items, with a total of 160 responses generated by ChatGPT (i.e., 5 runs of 1 block of 16 items). We first had these responses coded by a native speaker of English as DO, PO, or other sentences. Then we had the algorithm code the same set of model responses. There was a 100% match between the human and automatic coding (see osf.io/wkzr8 for the scripts and the coding test). We then used the algorithm to automatically code both prime and target completions as DO, PO, or "other" responses. Pairs in which either sentence was coded as "other" were removed from further analyses (22.8% and 20.3% of all the data respectively for ChatGPT and Vicuna).

*Syntax: syntactic ambiguity resolution*

The experiment began with the following instructions:

> *I will present you a small discourse containing several sentences, followed by a question about the discourse. Please only answer "yes" or "no" to the question according to preceding discourse. For instance, if you read "There was a tiger and a fox. The tiger ate the fox because it was hungry. Did the tiger eat the fox?", you should answer "Yes" to the question. If you read "There was a tiger and a fox. The tiger ate the fox because it was hungry. Did the fox escape from the tiger?", you should answer "No" to the question. Is that OK?*

The experiment had 32 target trials and 32 filler trials (osf.io/c28ur). A trial consisted of a context sentence and a target sentence, followed by a probe question (e.g., *There was a hunter and a poacher. The hunter killed the dangerous poacher with a rifle not long after sunset. Did the hunter use a rifle?*). ChatGPT/Vicuna was asked to answer "yes" or "no" to the probe question. We used automatic text extraction of "yes" or "no" from ChatGPT responses. If the method failed to extract "yes" or "no" from a response, a native speaker of English coded it manually and condition-blindly into "yes", "no", or "other". Responses coded as "other" were excluded from the analyses (22.8% and 20.3% of all the data respectively for ChatGPT and Vicuna).

*Meaning: implausible sentence interpretation*

The experiment began with the following instructions:

*I'd like to play a sentence comprehension game with you. I will give a sentence and a yes-or-no question regarding the sentence. Please simply answer "Yes" or "No" to the question. Shall we start?*

The stimuli were taken from Experiment 1.4 in Gibson et al. (2013), with 20 target trials and 40 filler trials (osf.io/2pktf). In a target trial, we presented ChatGPT/Vicuna with a sentence (plausible or implausible, in a DO or PO structure) together with a yes/no comprehension question (e.g., *The mother gave the candle the daughter. Did the daughter receive something/someone?*). We used automatic text extraction of "yes" or "no" from model responses; in trials where no "yes" or "no" was extracted, responses were manually inspected by a native speaker of English to determine if the response indicates a "yes" or "no" response; a trial was excluded if ChatGPT/Vicuna gave no clear indication of "yes" or "no" in its response (0.7% and 0.4% respectively for ChatGPT and Vicuna). A "yes"/"no" response was further coded as a literal interpretation of a target sentence (e.g., a "no" response to *The mother gave the candle the daughter. Did the daughter receive something/someone?*) or a nonliteral interpretation (e.g., a "yes" response to the above example).

*Meaning: semantic illusions*

The experiment began with the following instructions:

*I want you to answer some questions. Usually a one-word answer will be enough. If you don't know the answer, just say "don't know." You will occasionally encounter a question which has something wrong with it. For example, you might see the question: "When was President Gerald Ford forced to resign his office?" The thing that is wrong in this example is that Ford wasn't forced to resign. When you see a question like this, just say "'wrong." OK?*

The experiment contained 72 items, with 54 targets and 18 fillers (osf.io/r67f2); we divided these stimuli into two blocks (for ChatGPT). In a trial, we presented ChatGPT/Vicuna a question (e.g., *Snoopy is a black and white cat in what famous Charles Schulz comic strip?*), which it gave an answer or reported an error if it detected something wrong with the sentence. We coded whether a semantic illusion was detected by ChatGPT/Vicuna (by answering "wrong") or not (by giving any other answer). For Vicuna, 10 responses (out of 20,000) seemed to not relevant to the target question and were removed from the analyses.

*Discourse: implicit causality*

The experiment was run concurrently with the **structural priming** experiment in ChatGPT (but not in Vicuna). It began with the following instructions:

*I'd like to play a sentence completion game with you. I will provide a sentence preamble and I would like you to repeat the preamble and continue it into a full sentence.*

The experiment contained 32 target preambles (adapted from Fukumura & van Gompel, 2010) and 96 filler preambles, 64 of which were target stimuli from the structural priming experiment (osf.io/k3cfv); these stimuli were divided into two blocks in ChatGPT (but not in Vicuna). In a target trial, we presented ChatGPT/Vicuna with a sentence preamble in the format of subject-verb-object followed by *because* (e.g., *Gary scared Anna because ...*); the subject and object were personal names that differed in gender (with name gender counter-balanced between the subject and the object across items). ChatGPT/Vicuna repeated and completed the preamble (e.g., *Gary scared Anna because he jumped out from behind a tree and yelled "boo!"*). As in the ***sound-gender association*** experiment, we used automatic text extraction (*he/him/his/* vs *she/her/hers* following *because*) to code the completion as referring to the subject or the object. For responses where automatic text extraction failed to extract the pronouns or extracted multiple pronouns that differed in gender, two native English speakers independently and condition-blindly coded those items, with a third native English speaker resolving any discrepancies between the first two coders. Responses that included no pronouns, pronouns of different genders, or were otherwise ambiguous in terms of subject/object reference were coded as "other" (17% and 5% for ChatGPT and Vicuna respectively) and removed from further analyses.

*Discourse: drawing inferences*

The experiment began with the following instructions:

*I will present you with sentences and ask a yes or no question about those sentences. Please respond only with "yes", "no", or "don't know". Is that OK?*

The experiment contained 48 items (24 targets and 24 fillers; osf.io/e3wxc). A filler item comprised two sentences and a yes/no question (e.g., *While swimming in the shallow water near the rocks, Sharon cut her foot on a piece of glass. She had been looking for the watch that she misplaced while sitting on the rocks. Did she cut her foot?*). For target items, the question should elicit a "yes" response if inferences were made but a "no" response if no inference was made. We used automatic text extraction to extract the "yes" and "no" answers; a native speaker manually inspected a response if no "yes" or "no" response was detected. When a response indicated a "don't know" response (42% and 44% for ChatGPT and Vicuna respectively), it was excluded from further analyses.

*Interlocutor sensitivity: word meaning access*

The experiment began with a self-introduction of the simulated interlocutor (BE interlocutor or AE interlocutor).

*(BE interlocutor) Hi, I am a British English speaker. I am from the UK. I am now living in London and studying for a BA degree at King's College London.*

*(AE interlocutor) Hi, I am an American English speaker. I am from the USA. I am now living in New York and studying for a BA degree at the City University of New York.*

After ChatGPT/Vicuna responded to the self-introduction, we gave the following instructions for the experimental task:

> *I'd like to play a word association game with you. I will give you a word, and you are to give ONE word or phrase that you think of at reading the word I gave. For example, if I say "milk", you can say "cow" or "breakfast". I will give you the first word. Shall we start?*

The experiment contained 56 trials, with 36 target words that have different meanings between BE and AE (e.g., *bonnet*, see osf.io/k2jgd) and 20 filler words that do not. A trial began with an interlocutor typing a word (e.g., *bonnet*) and ChatGPT/Vicuna gave an associate (e.g., "hat"). We filtered the data for unique responses to each target word and had two native speakers of English, who were provided with definitions of the BE and AE meanings of target words, to independently and condition-blindly code these unique responses as relating the BE meaning of the target word (e.g., "car" as relating to the vehicle meaning of *bonnet*), the AE meaning (e.g., "hat" as relating to the headdress meaning of *bonnet*), or some other meaning. Any disagreement in coding (15.5% of all unique responses) was resolved by a third coder (also a native speaker of English). Trials where the associate related to "other" meanings or the response did not provide an associate (12% and 40% for ChatGPT and Vicuna respectively) were discarded from further analyses.

*Interlocutor sensitivity: lexical retrieval*

The experiment began with a self-introduction of the simulated interlocutors (BE interlocutor vs. AE interlocutor), using the same wording as in the ***Interlocutor sensitivity: word meaning access*** experiment. Then we gave instructions of the experimental task:

> *I'd like to play a word puzzle game with you. I will give you a definition and you are to supply the word/phrase that is defined. For example, if the definition is "an electronic device for storing and processing data, typically in binary form", you can say "computer". I will give you the first definition. Please only give me the defined word/phrase. Shall we start?*

The experiment contained 56 definitions, half of which were target definitions for which BE and AE have different lexical expressions (e.g., *potatoes deep-fried in thin strips* defines *chips* in BE but *French fries* in AE; see osf.io/28vt4). A trial began with the interlocutor typing a definition (e.g., *potatoes deep-fried in thin strips*) and ChatGPT/Vicuna giving the defined word/phrase (e.g., *French fries*). We filtered the data for unique responses for each definition and had two coders (native speakers of English) code these responses independently and condition-blindly as a BE expression, an AE expression, or an "other" expression, in reference to the BE/AE expressions associated with each definition. Variants of the reference BE/AE expressions (e.g., "economy class" instead of "economy", "chip" instead of "chips") were accepted as BE or AE expressions. Words/phrases that did not go with the reference expressions were coded as "other". Any disagreement in coding (5.1% of all unique responses) was resolved by a third coder (also a native speaker of English and again in a condition-blind manner). Trials with "other" expressions (5% and 21% for ChatGPT and Vicuna respectively) were discarded from further analyses.

**ACKNOWLEDGMENTS**

We thank Steven Langsford and Max Dunn for their help in data coding, and Ping Nie, Chi Fong Wong, Harvey Zhuang Qiu, and Steven Langsford for comments on the manuscript.

# REFERENCES

Altmann, G., & Steedman, M. (1988). Interaction with context during human sentence processing. *Cognition*, *30*(3), 191–238. https://doi.org/10.1016/0010-0277(88)90020-0

Aher, G. V., Arriaga, R. I., & Kalai, A. T. (2023, July). Using large language models to simulate multiple humans and replicate human subject studies. *Proceedings of the 40th International Conference on Machine Learning, PMLR 202:337-371*. https://proceedings.mlr.press/v202/aher23a.html

Argyle, L. P., Busby, E. C., Fulda, N., Gubler, J. R., Rytting, C., & Wingate, D. (2023). Out of one, many: Using language models to simulate human samples. *Political Analysis, 31*(3), 337-351. https://doi.org/10.1017/pan.2023.2

Bender, E. M., & Koller, A. (2020). Climbing towards NLU: On meaning, form, and understanding in the age of data. *Proceedings of the 58th Annual Meeting of the Association for Computational Linguistics*, 5185–5198. https://aclanthology.org/2020.acl-main.463

Bock, J. K. (1986). Syntactic persistence in language production. *Cognitive Psychology*, *18*(3), 355–387. https://doi.org/10.1016/0010-0285(86)90004-6

Brown, R., & Fish, D. (1983). The psychological causality implicit in language. *Cognition*, *14*(3), 237–273. https://doi.org/10.1016/0010-0277(83)90006-9

Brown, T., Mann, B., Ryder, N., Subbiah, M., Kaplan, J. D., Dhariwal, P., Neelakantan, A., Shyam, P., Sastry, G., Askell, A., Agarwal, S., Herbert-Voss, A., Krueger, G., Henighan, T., Child, R., Ramesh, A., Ziegler, D., Wu, J., Winter, C., … Amodei, D. (2020). Language models are few-shot learners. *Proceedings of the 34th International Conference on Neural Information Processing Systems*, *33*, 1877–1901. https://papers.nips.cc/paper/2020/hash/1457c0d6bfcb4967418bfb8ac142f64a-Abstract.html

Cai, Z. G. (2022). Interlocutor modelling in comprehending speech from interleaved interlocutors of different dialectic backgrounds. *Psychonomic Bulletin & Review*, *29*(3), 1026–1034. https://doi.org/10.3758/s13423-022-02055-7

Cai, Z. G., Dunn, M. S., & Branigan, H. P. (accepted in principle). How do speakers tailor lexical choices according to their interlocutor's accent? *Journal of Experimental Psychology: Learning, Memory, and Cognition*. https://osf.io/b3fcm/

Cai, Z. G., Gilbert, R. A., Davis, M. H., Gaskell, M. G., Farrar, L., Adler, S., & Rodd, J. M. (2017). Accent modulates access to word meaning: Evidence for a speaker-model account of spoken word recognition. *Cognitive Psychology*, *98*, 73–101. https://doi.org/10.1016/j.cogpsych.2017.08.003

Cai, Z. G., & Zhao, N. (2019). The sound of gender: Inferring the gender of names in a foreign language. *Journal of Cultural Cognitive Science*, *3*, 63–73. https://doi.org/10.1007/s41809-019-00028-2

Cai, Z. G., Zhao, N., & Pickering, M. J. (2022). How do people interpret implausible sentences? *Cognition*, *225*, 105101. https://doi.org/10.1016/j.cognition.2022.105101

Cai, Z. G., & Zhao, N. (2024). Structural priming: An experimental paradigm for mapping linguistic representations. *Language and Linguistics Compass, 18*(2), e12507. https://doi.org/10.1111/lnc3.12507

Cassani, G., Chuang, Y.-Y., & Baayen, R. H. (2020). On the semantics of nonwords and their lexical category. *Journal of Experimental Psychology: Learning, Memory, and Cognition*, *46*(4), 621. https://doi.org/10.1037/xlm0000747

Cassidy, K. W., Kelly, M. H., & Sharoni, L. J. (1999). Inferring gender from name phonology. *Journal of Experimental Psychology: General*, *128*(3), 362–381. https://doi.org/10.1037/0096-3445.128.3.362

Caucheteux, C., & King, J.-R. (2022). Brains and algorithms partially converge in natural language processing. *Communications Biology*, *5*(1), 134. https://doi.org/10.1038/s42003-022-03036-1

Chiang, W.-L., Li, Z., Lin, Z., Sheng, Y., Wu, Z., Zhang, H., Zheng, L., Zhuang, S., Zhuang, Y., & Gonzalez, J. E. (2023). Vicuna: An open-source chatbot impressing gpt-4 with 90%* chatgpt quality. https://lmsys.org/blog/2023-03-30-vicuna/

Chomsky, N. (2000). *New horizons in the study of language and mind*. Cambridge University Press.

Chomsky, N., Roberts, I., & Watumull, J. (2023, March 8). *Noam Chomsky: The false promise of ChatGPT*. The New York Times. https://archive.is/AgWkn#selection-317.0-317.13

Christiano, P. F., Leike, J., Brown, T., Martic, M., Legg, S., & Amodei, D. (2017). Deep reinforcement learning from human preferences. *Advances in Neural Information Processing Systems, 30*. https://proceedings.neurips.cc/paper_files/paper/2017/hash/d5e2c0adad503c91f91df240d0cd4e49-Abstract.html

Cowan, B. R., Doyle, P., Edwards, J., Garaialde, D., Hayes-Brady, A., Branigan, H. P., Cabral, J., & Clark, L. (2019). What's in an accent? The impact of accented synthetic speech on lexical choice in human-machine dialogue. *Proceedings of the 1st International Conference on Conversational User Interfaces*, 1–8. https://doi.org/10.1145/3342775.3342786

Cutler, A., McQueen, J., & Robinson, K. (1990). Elizabeth and John: Sound patterns of men's and women's names. *Journal of Linguistics*, *26*(2), 471–482. https://doi.org/doi:10.1017/S0022226700014754

Ćwiek, A., Fuchs, S., Draxler, C., Asu, E. L., Dediu, D., Hiovain, K., Kawahara, S., Koutalidis, S., Krifka, M., & Lippus, P. (2022). The bouba/kiki effect is robust across cultures and writing systems. *Philosophical Transactions of the Royal Society B*, *377*(1841), 20200390. https://doi.org/10.1098/rstb.2020.0390

Dentella, V., Günther, F., & Leivada, E. (2023). Systematic testing of three Language Models reveals low language accuracy, absence of response stability, and a yes-response bias. *Proceedings of the National Academy of Sciences, 120*(51), e2309583120.

Devlin, J., Chang, M.-W., Lee, K., & Toutanova, K. (2019). *BERT: Pre-training of Deep Bidirectional Transformers for Language Understanding*. arXiv. https://doi.org/10.48550/arXiv.1810.04805

Dunn, M. S., & Cai, Z. G. (2023). *Word length affects language production in (non)predictive contexts but not language comprehension* [poster presentation]. AMLaP Asia, Hong Kong.

Erickson, T. D., & Mattson, M. E. (1981). From words to meaning: A semantic illusion. *Journal of Verbal Learning and Verbal Behavior*, *20*(5), 540–551. https://doi.org/10.1016/S0022-5371(81)90165-1

Ethayarajh, K. (2019). *How contextual are contextualized word representations? Comparing the geometry of BERT, ELMo, and GPT-2 embeddings*. arXiv. https://doi.org/10.48550/arXiv.1909.00512

Fukumura, K., & van Gompel, R. P. G. (2010). Choosing anaphoric expressions: Do people take into account likelihood of reference? *Journal of Memory and Language*, *62*(1), 52–66. https://doi.org/10.1016/j.jml.2009.09.001

Garvey, C., & Caramazza, A. (1974). Implicit causality in verbs. *Linguistic Inquiry*, *5*(3), 459–464.

Gatti, D., Marelli, M., & Rinaldi, L. (2022). Out-of-vocabulary but not meaningless: Evidence for semantic-priming effects in pseudoword processing. *Journal of Experimental Psychology: General*, *152*(3), 851–863. https://doi.org/10.1037/xge0001304

Gibson, E., Bergen, L., & Piantadosi, S. T. (2013). Rational integration of noisy evidence and prior semantic expectations in sentence interpretation. *Proceedings of the National Academy of Sciences*, *110*(20), 8051–8056. https://doi.org/10.1073/pnas.1216438110

Gilbert, R. A., & Rodd, J. M. (2022). Dominance norms and data for spoken ambiguous words in British English. *Journal of Cognition*, *5(1)*, 4. https://doi.org/10.5334/joc.194

Goldberg, Y. (2019). *Assessing BERT's syntactic abilities*. https://doi.org/10.48550/arXiv.1901.05287

Gulordava, K., Bojanowski, P., Grave, E., Linzen, T., & Baroni, M. (2018). *Colorless green recurrent networks dream hierarchically*. arXiv. https://arxiv.org/abs/1803.11138

Hannon, B., & Daneman, M. (2001). Susceptibility to semantic illusions: An individual-differences perspective. *Memory & Cognition*, *29*(3), 449–461. https://doi.org/10.3758/BF03196396

Haslett, D. A., & Cai, Z. G. (2023). Similar-sounding words flesh out fuzzy meanings. *Journal of Experimental Psychology: General*. *152*(8), 2359–2368. https://doi.org/10.1037/xge0001409

Hughes, A. (2023). *ChatGPT: Everything you need to know about OpenAI's GPT-4 tool*. BBC Science Focus Magazine. https://www.sciencefocus.com/future-technology/gpt-3/

Jain, S., Vo, V. A., Wehbe, L., & Huth, A. G. (2023). Computational language modeling and the promise of in silico experimentation. *Neurobiology of Language*, 1-27. https://doi.org/10.1162/nol_a_00101

Ji, Z., Lee, N., Frieske, R., Yu, T., Su, D., Xu, Y., ... & Fung, P. (2023). Survey of hallucination in natural language generation. *ACM Computing Surveys, 55*(12), 1-38. https://doi.org/10.1145/3571730

Kaushal, A., & Mahowald, K. (2022). *What do tokens know about their characters and how do they know it?* arXiv. https://doi.org/10.48550/arXiv.2206.02608

Köhler, W. (1929). *Gestalt Psychology*. Liveright.

Levy, R., Bicknell, K., Slattery, T., & Rayner, K. (2009). Eye movement evidence that readers maintain and act on uncertainty about past linguistic input. *Proceedings of the National Academy of Sciences*, *106*(50), 21086–21090. https://doi.org/10.1073/pnas.0907664106

Linzen, T., & Baroni, M. (2021). Syntactic Structure from Deep Learning. *Annual Review of Linguistics*, *7*(1), 195–212. https://doi.org/10.1146/annurev-linguistics-032020-051035

Liu, Y., Ott, M., Goyal, N., Du, J., Joshi, M., Chen, D., Levy, O., Lewis, M., Zettlemoyer, L., & Stoyanov, V. (2019). *RoBERTa: A robustly optimized BERT pretraining approach*. arXiv. https://doi.org/10.48550/arXiv.1907.11692

Mahowald, K. (2023). *A Discerning Several Thousand Judgments: GPT-3 Rates the Article + Adjective + Numeral + Noun Construction*. arXiv. https://doi.org/10.48550/arXiv.2301.12564

Mahowald, K., Fedorenko, E., Piantadosi, S. T., & Gibson, E. (2013). Info/information theory: Speakers choose shorter words in predictive contexts. *Cognition*, *126*(2), 313–318. https://doi.org/10.1016/j.cognition.2012.09.010

Mahowald, K., Ivanova, A. A., Blank, I. A., Kanwisher, N., Tenenbaum, J. B., & Fedorenko, E. (2023). *Dissociating language and thought in large language models: A cognitive perspective*. arXiv. http://arxiv.org/abs/2301.06627

Marvin, R., & Linzen, T. (2018). *Targeted syntactic evaluation of language models*. arXiv. https://arxiv.org/abs/1808.09031

McCoy, R. T., Pavlick, E., & Linzen, T. (2019). *Right for the wrong reasons: Diagnosing syntactic heuristics in natural language inference*. arXiv. https://doi.org/10.48550/arXiv.1902.01007

McKoon, G., & Ratcliff, R. (1986). Inferences about predictable events. Journal of Experimental Psychology: *Learning, Memory, and Cognition*, 12, 82–91. https://doi.org/10.1037/0278-7393.12.1.82

Michaelov, J., Arnett, C., Chang, T., & Bergen, B. (2023). Structural priming demonstrates abstract grammatical representations in multilingual language models. *Proceedings of the 2023 Conference on Empirical Methods in Natural Language Processing*, 3703–3720. https://aclanthology.org/2023.emnlp-main.227/

OpenAI. (2022, November 30). *Introducing ChatGPT*. https://openai.com/blog/chatgpt

Ouyang, L., Wu, J., Jiang, X., Almeida, D., Wainwright, C., Mishkin, P., Zhang, C., Agarwal, S., Slama, K., & Ray, A. (2022). Training language models to follow instructions with human feedback. *Advances in Neural Information Processing Systems*, *35*. https://proceedings.neurips.cc/paper_files/paper/2022/hash/b1efde53be364a73914f58805a001731-Abstract-Conference.html

Peters, M. E., Neumann, M., Iyyer, M., Gardner, M., Clark, C., Lee, K., & Zettlemoyer, L. (2018). *Deep contextualized word representations*. arXiv. https://doi.org/10.48550/arXiv.1802.05365

Piantadosi, S. (2023). *Modern language models refute Chomsky's approach to language*. Lingbuzz Preprint, 7180. https://lingbuzz.net/lingbuzz/007180/v1.pdf

Piantadosi, S. T., Tily, H., & Gibson, E. (2011). Word lengths are optimized for efficient communication. *Proceedings of the National Academy of Sciences*, *108*(9), 3526–3529. https://doi.org/10.1073/pnas.1012551108

Pickering, M. J., & Branigan, H. P. (1998). The representation of verbs: Evidence from syntactic priming in language production. *Journal of Memory and Language*, *39*(4), 633–651. https://doi.org/10.1006/jmla.1998.2592

Prasad, G., Van Schijndel, M., & Linzen, T. (2019). *Using priming to uncover the organization of syntactic representations in neural language models*. arXiv. https://doi.org/10.48550/arXiv.1909.10579

Qiu, Z., Duan, X., & Cai, Z. G. (2023). *Pragmatic Implicature Processing in ChatGPT*. PsyArXiv. https://doi.org/10.31234/osf.io/qtbh9.

Qiu, Z., Duan, X., & Cai, Z. G. (2024). *Grammaticality Representation in ChatGPT as Compared to Linguists and Laypeople*. Retrieved from osf.io/preprints/psyarxiv/r9zdh

Radford, A., Narasimhan, K., Salimans, T., & Sutskever, I. (2018). Improving language understanding by generative pre-training. https://www.mikecaptain.com/resources/pdf/GPT-1.pdf

Radford, A., Wu, J., Child, R., Luan, D., Amodei, D., & Sutskever, I. (2019). Language models are unsupervised multitask learners. https://insightcivic.s3.us-east-1.amazonaws.com/language-models.pdf

Rayner, K., Carlson, M., & Frazier, L. (1983). The interaction of syntax and semantics during sentence processing: Eye movements in the analysis of semantically biased sentences. *Journal of Verbal Learning and Verbal Behavior*, *22*(3), 358–374. https://doi.org/10.1016/S0022-5371(83)90236-0

Reder, L. M., & Kusbit, G. W. (1991). Locus of the Moses illusion: Imperfect encoding, retrieval, or match? *Journal of Memory and Language*, *30*(4), 385–406. https://doi.org/10.1016/0749-596X(91)90013-A

Rodd, J. M., Lopez Cutrin, B., Kirsch, H., Millar, A., & Davis, M. H. (2013). Long-term priming of the meanings of ambiguous words. *Journal of Memory and Language*, *68*(2), 180–198. https://doi.org/10.1016/j.jml.2012.08.002

Schrimpf, M., Blank, I. A., Tuckute, G., Kauf, C., Hosseini, E. A., Kanwisher, N., Tenenbaum, J. B., & Fedorenko, E. (2021). The neural architecture of language: Integrative modeling converges on predictive processing. *Proceedings of the National Academy of Sciences*, *118*(45), e2105646118. https://doi.org/10.1073/pnas.2105646118

Sidhu, D. M., & Pexman, P. M. (2017). A prime example of the Maluma/Takete effect? Testing for sound symbolic priming. *Cognitive Science*, *41*(7), 1958–1987. https://doi.org/10.1111/cogs.12438

Sinclair, A., Jumelet, J., Zuidema, W., & Fernández, R. (2022). Structural persistence in language models: Priming as a window into abstract language representations. *Transactions of the Association for Computational Linguistics, 10*, 1031–1050. https://doi.org/10.1162/tacl_a_00504

Singer, M., & Spear, J. (2015). Phantom recollection of bridging and elaborative inferences. *Discourse Processes*, *52*(5–6), 356–375. https://doi.org/10.1080/0163853X.2015.1029858

Sun, S., Zhang, Y., Yan, J., Gao, Y., Ong, D., Chen, B., & Su, J. (2023, October 16). Battle of the Large Language Models: Dolly vs LLaMA vs Vicuna vs Guanaco vs Bard vs ChatGPT -- A Text-to-SQL Parsing Comparison. *Findings of the Association for Computational Linguistics: EMNLP 2023*. 11225–11238. https://doi.org/10.48550/arXiv.2310.10190

Tenney, I., Das, D., & Pavlick, E. (2019). *BERT rediscovers the classical NLP pipeline*. arXiv. https://doi.org/10.48550/arXiv.1905.05950

Van Berkum, J. J. A., van den Brink, D., Tesink, C. M. J. Y., Kos, M., & Hagoort, P. (2008). The neural integration of speaker and message. *Journal of Cognitive Neuroscience*, *20*(4), 580–591. https://doi.org/10.1162/jocn.2008.20054

van Gompel, R. P. G., Pickering, M. J., & Traxler, M. J. (2001). Reanalysis in sentence processing: Evidence against current constraint-based and two-stage models. *Journal of Memory and Language*, *45*(2), 225–258. https://doi.org/10.1006/jmla.2001.2773

van Gompel, R. P., Wakeford, L. J., & Kantola, L. (2023). No looking back: the effects of visual cues on the lexical boost in structural priming. *Language, Cognition and Neuroscience*, 38(1), 1-10.

Vaswani, A., Shazeer, N., Parmar, N., Uszkoreit, J., Jones, L., Gomez, A. N., Kaiser, Ł., & Polosukhin, I. (2017). Attention is all you need. *Advances in Neural Information Processing Systems*, *30*. https://proceedings.neurips.cc/paper/7181-attention-is-all

Wei, J., Tay, Y., Bommasani, R., Raffel, C., Zoph, B., Borgeaud, S., Yogatama, D., Bosma, M., Zhou, D., Metzler, D., Chi, E. H., Hashimoto, T., Vinyals, O., Liang, P., Dean, J., & Fedus, W. (2022). *Emergent abilities of large language models*. arXiv. https://doi.org/10.48550/arXiv.2206.07682

Xu, Z., Jain, S., & Kankanhalli, M. (2024). *Hallucination is inevitable: An innate limitation of large language models*. arXiv. https://doi.org/10.48550/arXiv.2401.11817

**SUPPLEMENTAL INFORMATION**

We provided exploratory analyses (preregistered or non-preregistered) here; preregistered exploratory analyses can also be viewed in the preregistrations (osf.io/vu2h3/registrations).

*Sounds: sound-shape association*

In a non-preregistered analysis, we tested the possibility that an LLM might have been trained on the papers (or their abstracts) on which our experiments were based and associated a psycholinguistic effect with the exemplar stimuli used in the paper/abstract to illustrate the psycholinguistic effect. If this is the case, we should expect the effect to disappear if we removed the exemplar items from the analyses. Thus, in this experiment, we removed 6 exemplar items (e.g., *maluma*, *takete*), leaving the remaining 14 items for analyses. We observed that excluding the exemplar items did not affect the pattern of results, with round-sounding words still being judged to be round in shape more often than spike-sounding words in both ChatGPT (0.80 vs. 0.58, $\beta = 1.58$, $SE = 0.36$, $z = 4.37$, $p < .001$) and Vicuna (0.39 vs. 0.31, $\beta = 0.35$, $SE = 0.15$, $z = 2.36$, $p = .018$).

In another non-preregistered analysis, we conducted a post-test to see whether ChatGPT identified any of the novel words as English words. It identified *maluma* as an English word almost half the time (8 of 20 trials), so we conducted the same LME analyses as in the main text but while excluding that item. The effect was almost the same as when *maluma* was included: ChatGPT assigned round-sounding novel words to round shapes more often than it assigned spiky-sounding novel words to round shapes (0.79 vs. 0.49, $\beta = 2.03$, $SE = 0.36$, $z = 5.65$, $p < .001$); so did Vicuna (0.39 vs. 0.32, $\beta = 0.28$, $SE = 0.12$, $z = 2.36$, $p = .018$).

Following our preregistered exploratory analysis, we calculated the proportion of "round" responses for each item and compared that value to the proportion of "round" responses per item by human participants, as reported by Sidhu & Pexman (2017). We found a significant 0.85 correlation between ChatGPT responses and human responses ($t(16) = 6.53$, $p < .001$) and a nonsignificant 0.18 correlation between Vicuna responses and human responses ($t(16) = 0.75$, $p = .463$).

*Sounds: sound-gender association*

We conducted a non-preregistered analysis by removing 1 exemplar item (i.e., *Corla/Colark*), leaving 15 items in the analysis. The pattern of effects still held, with more use of feminine pronouns to refer to a name ending with a vowel than to one ending with a consonant in both ChatGPT (0.74 vs. 0.23, $\beta = 4.79$, $SE = 1.25$, $z = 3.84$, $p < .001$) and Vicuna (0.39 vs. 0.02, $\beta = 5.40$, $SE = 1.23$, $z = 4.41$, $p < .001$).

*Words: word length and predictivity*

We conducted a non-preregistered analysis by removing 1 exemplar item (i.e., *math/mathematics*), leaving 39 items in the analysis. The exclusion did not change the pattern of results, with no significant difference between the predictive and neutral contexts in both ChatGPT (0.24 vs. 0.19, $\beta = 0.29$, $SE = 0.22$, $z = 1.32$, $p = .188$) and Vicuna (0.31 vs. 0.31, $\beta = -0.16$, $SE = 0.20$, $z = -0.77$, $p = .439$).

We also conducted a non-preregistered exploratory analysis comparing trial-level data between language models (ChatGPT/Vicuna) and human participants (from Mahowald et al., 2013), treating context and participant group (humans = -0.5, ChatGPT/Vicuna = 0.5) as interacting predictors. We observed a significant difference between ChatGPT/Vicuna and humans, with LLMs being less likely to choose the short word than human participants (ChatGPT vs. humans: *β* = -3.14, *SE* = 0.22, *z* = -13.98, *p* < .001; Vicuna vs. humans: *β* = -1.86, *SE* = 0.24, *z* = -7.61, *p* < .001; see also Fig 1 bottom left). There was also an effect of context in the ChatGPT-human comparison, with the short word chosen more often in a predictive than neutral context (*β* = 0.44, *SE* = 0.16, *z* = 2.83, *p* < .005) but there was no such an effect in the Vicuna-human comparison (*β* = 0.15, *SE* = 0.12, *z* = 1.24, *p* = .215). The effect of context was similar between ChatGPT and humans, as indicated by the lack of an interaction between group and context (*β* = -0.20, *SE* = 0.20, *z* = -0.96, *p* = .336), but the effect of context was larger in humans than in Vicuna, as indicated by the significant interaction between group and context (*β* = -0.60, *SE* = 0.22, *z* = -2.76, *p* = .006).

*Words: word meaning priming*

We also conducted a non-preregistered analysis by removing 14 exemplar items (e.g., *post*), leaving 25 items in the analysis. In both models, there was no significant difference in meaning access between a synonym prime and no prime (ChatGPT: 0.38 vs. 0.33, *β* = 0.36, *SE* = 0.19, *z* = 1.90, *p* = .057; Vicuna: 0.19 vs. 0.15, *β* = 0.39, *SE* = 0.28, *z* = 1.40, *p* = .162); there was a significant word-meaning priming effect, with more access to the primed (subordinate) meaning following a word-meaning prime than following no prime (0.53 vs. 0.33, *β* = 2.47, *SE* = 0.30, *z* = 8.20, *p* < .001; Vicuna: 0.32 vs. 0.15, *β* = 3.33, *SE* = 0.50, *z* = 6.70, *p* < .001) and than following a synonym prime (ChatGPT: 0.53 vs. 0.38, *β* = 2.65, *SE* = 0.40, *z* = 6.58, *p* < .001; Vicuna: 0.32 vs. 0.19, *β* = 2.86, *SE* = 0.48, *z* = 5.91, *p* < .001).

Rodd et al. (2013, Experiment 3) also performed a secondary analysis where they removed any associate that is a morphological variant of a word in the prime sentence corresponding to an association trial; for example, if a participant gave *firm* or *accountant* as an associate to *post* following the prime sentence *The man accepted the post in the accountancy firm*, that trial was removed from the analysis. We initially preregistered this analysis but later changed to the main analysis in Rodd et al. (2013), as the removal method would lead to a lot of removals in the synonym prime condition, because the synonym could often be given as an associate to the target word (e.g., *job* as an associate of *post*). Nonetheless, we also followed the secondary analysis in Rodd et al. (2013) by excluding associates with the same lemma as any word in the corresponding prime sentence (e.g., we excluded *posting*, *firms*, or *accept* as associates of *post* following the word-meaning prime). Compared to the no-prime condition, the synonym prime led to less subordinate meaning access in ChatGPT (0.33 vs. 0.22, *β* = -0.79, *SE* = 0.32, *z* = -2.47, *p* = .013) but led to similar access in Vicuna (0.09 vs. 0.11, *β* = 0.44, *SE* = 0.30, *z* = 1.46, *p* = .146); critically, the word-meaning prime led to more subordinate meaning access than no prime (ChatGPT: 0.47 vs. 0.33, *β* = 1.88, *SE* = 0.37, *z* = 5.10, *p* < .001; Vicuna: 0.15 vs. 0.09, *β* = 2.79, *SE* = 0.51, *z* = 5.50, *p* < .001) and than the synonym prime (ChatGPT: 0.47 vs. 0.22, *β* = 2.65, *SE* = 0.40, *z* = 6.58, *p* < .001; Vicuna: 0.15 vs. 0.11, *β* = 2.86, *SE* = 0.50, *z* = 5.67, *p* < .001).

*Syntax: structural priming*

We conducted a non-preregistered analysis by removing 1 exemplar item, leaving 31 items in the analysis. The exclusion did not alter the pattern of results. For ChatGPT, there was a significant main effect of prime structure, with more PO responses following PO and DO primes (ChatGPT: 0.72 vs. 0.59, $\beta = 1.06$, $SE = 0.11$, $z = 9.67$, $p < .001$; Vicuna: 0.81 vs. 0.51, $\beta = 2.97$, $SE = 0.35$, $z = 8.49$, $p < .001$); there was no significant main effect of verb type, with similar PO responses when the prime and target had different verbs and when they had same verb (ChatGPT: 0.64 vs. 0.67, $\beta = -0.06$, $SE = 0.09$, $z = -0.63$, $p = .528$; Vicuna: 0.66 vs. 0.67, $\beta = -0.17$, $SE = 0.23$, $z = -0.75$, $p = .454$); there was a significant interaction, with a stronger structural priming effect when the verb was the same between the prime and target than when it was different (ChatGPT: 0.15 vs. 0.10 in priming effects, $\beta = 0.40$, $SE = 0.15$, $z = 2.63$, $p = .009$; Vicuna: 0.38 vs. 0.21 in priming effects, $\beta = 1.16$, $SE = 0.47$, $z = 2.49$, $p = .013$).

*Syntax: syntactic ambiguity resolution*

We conducted a non-preregistered analysis by removing 1 exemplar item (Example 5 in the main text), leaving 31 items in the analysis. The exclusion did not alter the pattern of results. There were more VP than NP attachments (ChatGPT: 0.94 vs. 0.06, $\beta = -9.35$, $SE = 0.74$, $z = -12.68$, $p < .001$; Vicuna: 0.63 vs. 0.37, $\beta = -1.33$, $SE = 0.16$, $z = -8.12$, $p < .001$). There was an effect of context in Vicuna, with more NP attachment interpretations following a multiple-referent context than following a single-referent context (0.38 vs. 0.36, $\beta = 0.20$, $SE = 0.10$, $z = 2.10$, $p = .036$) but not in ChatGPT (0.06 vs. 0.06, $\beta = -0.10$, $SE = 0.43$, $z = -0.23$, $p = .820$). There was an effect of question, with more NP attachment interpretations for an NP probe than for a VP probe (ChatGPT: 0.09 vs. 0.03, $\beta = 3.37$, $SE = 0.99$, $z = 3.42$, $p < .001$; Vicuna: 0.72 vs. 0.03, $\beta = 5.64$, $SE = 0.48$, $z = 11.76$, $p < .001$), and no interaction between context and probe (ChatGPT: $\beta = 0.19$, $SE = 0.70$, $z = 0.27$, $p = .785$; Vicuna: $\beta = -0.18$, $SE = 0.25$, $z = -0.71$, $p = .480$).

*Meaning: implausible sentence interpretation*

We conducted a non-preregistered analysis by removing 2 exemplar items (*The mother gave the daughter to the candle* and *The girl tossed the apple the boy*), leaving 18 items in the analysis. There was an effect of implausibility in ChatGPT, with more nonliteral interpretations for implausible than plausible sentences (0.75 vs. 0.02, $\beta = 11.59$, $SE = 0.69$, $z = 16.90$, $p < .001$) but not in Vicuna (0.49 vs. 0.37, $\beta = 1.99$, $SE = 1.30$, $z = 1.53$, $p = .126$). There was an effect of structure in ChatGPT, with more nonliteral interpretations for DO than PO sentences (0.48 vs. 0.29, $\beta = 1.59$, $SE = 0.45$, $z = 3.56$, $p < .001$) but not in Vicuna (0.45 vs. 0.41, $\beta = -0.10$, $SE = 0.38$, $z = -0.26$, $p = .794$). There was a significant interaction between plausibility and structure in ChatGPT, with the effect of plausibility being stronger in DO sentences than in PO sentences ($\beta = 2.55$, $SE = 0.76$, $z = 3.38$, $p < .001$) but not in Vicuna ($\beta = 1.24$, $SE = 0.66$, $z = 1.88$, $p = .060$). Analysing implausible sentences alone revealed an effect of structure, with more nonliteral interpretations for implausible DO than PO sentences in ChatGPT (0.92 vs. 0.57, $\beta = 3.13$, $SE = 0.68$, $z = 4.58$, $p < .001$) but not in Vicuna (0.52 vs. 0.46 $\beta = 0.51$, $SE = 0.54$, $z = 0.95$, $p = .342$).

In another non-preregistered analysis, we also compared trial-level data between ChatGPT/Vicuna and human participants (from Experiment 1.4 in Gibson et al., 2013) in the interpretation of implausible sentences (excluding plausible sentences). Compared to human participants, ChatGPT had more nonliteral interpretations of implausible sentences (0.45 vs. 0.74, $\beta = 2.08$, $SE = 0.44$, $z = 4.76$, $p < .001$), but Vicuna did not (0.45 vs. 0.50, $\beta = 0.45$, $SE = 0.54$, $z = 0.83$, $p = .410$). There is an effect of structure, with more nonliteral interpretations for implausible DOs than implausible POs in both the ChatGPT/human comparison (0.90 vs. 0.55, $\beta$

= 2.15, *SE* = 0.43, *z* = 5.03, *p* < .001) and the Vicuna/human comparison (0.54 vs. 0.46, *β* = 0.68, *SE* = 0.31, *z* = 2.16, *p* = .031). The interaction between group and structure was significant in the ChatGPT/human comparison, suggesting that the effect of structure was larger in ChatGPT than in humans (*β* = 2.75, *SE* = 0.75, *z* = 3.68, *p* < .001), but the interaction was not significant in the Vicuna/human comparison (*β* = 0.18, *SE* = 0.55, *z* = 0.33, *p* = .739).

*Meaning: semantic illusions*

We conducted a non-preregistered analysis by removing 2 exemplar items ("What board game includes bishops/cardinals/monks, rooks, pawns, knights, kings, and queens?" and "What passenger liner was tragically sunk by an iceberg in the Atlantic/Pacific/Indian Ocean?"), leaving 52 items in the analysis. In ChatGPT, compared to the baseline, there were more error reports in the strong imposter conditions (0.00 vs. 0.14, *β* = 14.40, *SE* = 1.15, *z* = 12.55, *p* < .001) and in the weak imposter condition (0.00 vs. 0.17, *β* = 15.24, *SE* = 1.15, *z* = 13.27, *p* < .001); there was no statistical difference in error reports between the two imposter conditions (*β* = 1.33, *SE* = 0.83, *z* = 1.60, *p* = .109). In Vicuna, there was no statistical difference in error reports between the baseline and the strong imposter condition (0.002 vs. 0.022, *β* = -2.78, *SE* = 1.62, *z* = -1.72, *p* = .085) or between the baseline and the weak imposter condition (0.002 vs. 0.018, *β* = 1.00, *SE* = 1.30, *z* = 0.77, *p* = .445); the weak imposter condition led to more error reports than the strong imposter condition (*β* = 3.90, *SE* = 1.16, *z* = 3.36, *p* < .001), though numerically there was a lower error report rate in the weak than strong imposter condition (0.022 vs. 0.017).

*Discourse: implicit causality*

We conducted a non-preregistered analysis by removing 3 exemplar items (*Gary scared Anna because he was wearing a mask and making strange noises*, *Toby impressed Susie because he got a perfect score on the math exam*, and *Brian impressed Janet because of his exceptional intelligence and charming personality*), leaving 29 items in the analysis. The exclusion did not alter the pattern of results: more completions with a pronoun referring to the object following an experiencer-stimulus verb than following a stimulus-experiencer verb in ChatGPT (0.95 vs. 0.00, *β* = 13.82, *SE* = 0.94, *z* = 14.69, *p* < .001) and also in Vicuna (0.91 vs. 0.01, *β* = 14.37, *SE* = 1.51, *z* = 9.54, *p* < .001).

*Discourse: drawing inferences*

We conducted a non-preregistered analysis by removing 1 exemplar item (the example in (9) in the main text), leaving 23 items in the analysis. The exclusion did not change the results pattern. In both models, compared to the explicit condition, there were fewer "yes" responses in the bridging condition (ChatGPT: 0.49 vs. 0.95, *β* = -5.05, *SE* = 0.10, *z* = -50.06, *p* < .001; Vicuna: 0.24 vs. 0.79, *β* = -4.37, *SE* = 0.52, *z* = -8.33, *p* < .001) and in the elaborative condition (ChatGPT: 0.23 vs. 0.95, *β* = -7.40, *SE* = 0.12, *z* = -62.59, *p* < .001; Vicuna: 0.20 vs. 0.79, *β* = -4.43, *SE* = 0.43, *z* = -10.22, *p* < .001). Critically, ChatGPT made fewer "yes" responses in the elaborative than bridging condition (0.23 vs. 0.49, *β* = -2.94, *SE* = 0.60, *z* = -4.89, *p* < .001), whereas Vicuna made similar "yes" responses between the bridging and elaborative conditions (0.24 vs. 0.20, *β* = -0.06, *SE* = 0.44, *z* = -0.13, *p* = .900).

*Interlocutor sensitivity: word meaning access*

We conducted a non-preregistered analysis by removing 13 exemplar items (e.g., "*bonnet*"), leaving 23 items in the analysis. The exclusion did not alter the pattern of results.

There was more access to the AE meaning with an AE interlocutor than a BE interlocutor in both ChatGPT (0.46 vs. 0.36, $\beta = 1.84$, $SE = 0.25$, $z = 7.28$, $p < .001$) and in Vicuna (0.62 vs. 0.33, $\beta = 2.80$, $SE = 0.54$, $z = 5.15$, $p < .001$).

Following the preregistered exploratory analysis, we also included (log) trial order (i.e., the log order in which a target trial was presented, among both targets and fillers, to ChatGPT in an experimental run) (Note that the Vicuna experiment had one trial per run so there was no trial order). This analysis was to see if the interlocutor sensitivity (if any) varies over time. Thus, the LME model included interlocutor and (log) trial order as interacting predictors. We observed a significant interlocutor effect ($\beta = 2.01$, $SE = 0.25$, $z = 7.91$, $p < .001$), with more access to AE meanings for an AE than BE interlocutor, and a significant effect of trial order ($\beta = -0.49$, $SE = 0.15$, $z = -3.20$, $p = .001$), with decreasing AE meaning access over time. Importantly, we also observed a significant interaction between interlocutor and (log) trial order ($\beta = -0.54$, $SE = 0.17$, $z = -3.14$, $p = .002$), showing that the interlocutor effect decreased over time. Such a decrease of interlocutor sensitivity is not observed in human experiments (e.g., Cai et al., 2017) and might be due to the attenuating contextual influence (i.e., the interlocutor dialectal background) over time in ChatGPT.

In a non-preregistered analysis, we also compared trial-level data between ChatGPT/Vicuna and human participants (pooled from Experiment 1 of Cai et al., 2017) and the blocked condition of Experiment 1 of Cai (2022). There was no effect of participant group (ChatGPT: $\beta = 0.53$, $SE = 0.91$, $z = 0.58$, $p = .560$; Vicuna: $\beta = 0.65$, $SE = 0.68$, $z = 0.96$, $p = .338$), with a similar proportion of AE meaning access for ChatGPT/Vicuna and human participants (see Fig 3 bottom left). There was an interlocutor effect (ChatGPT: $\beta = 1.13$, $SE = 0.14$, $z = 8.28$, $p < .001$; Vicuna: $\beta = 1.59$, $SE = 0.27$, $z = 5.97$, $p < .001$), with more access to AE meanings for words produced by an AE interlocutor than by a BE interlocutor. There was also an interaction between group and interlocutor (ChatGPT: $\beta = 1.34$, $SE = 0.28$, $z = 4.70$, $p < .001$; Vicuna: $\beta = 2.26$, $SE = 0.58$, $z = 3.92$, $p < .001$), which suggests that ChatGPT/Vicuna was more sensitive to an interlocutor's dialectal background in word meaning access than human participants were (however, it should be noted that ChatGPT/Vicuna was explicitly told about an interlocutor's dialectic background, whereas human participants inferred their dialectal background via their accent).

*Interlocutor sensitivity: lexical retrieval*

Note that that the human study on which this experiment was based was not published at the time of experiment so we did not conduct any analysis excluding exemplar items.

Following the preregistered exploratory analysis, we also included (log) trial order (i.e., the log order in which a target trial was presented, among both targets and fillers, to ChatGPT in an experimental run). In an LME model with interlocutor and (log) trial order as interacting predictors, we observed an interlocutor effect ($\beta = 4.21$, $SE = 1.76$, $z = 2.39$, $p = .017$; with more AE meaning access for words from an AE interlocutor than from a BE interlocutor), a trial order effect ($\beta = 1.51$, $SE = 0.60$, $z = 2.54$, $p = .011$; with increasing AE expressions over time), and an interaction between interlocutor and trial order ($\beta = -0.18$, $SE = 0.09$, $z = -2.17$, $p = .030$; with a decreasing interlocutor effect over time).

In a non-preregistered analysis, we also compared trial-level data between ChatGPT and human participants (from the pilot experiment of Cai et al., accepted in principle) and between Vicuna and human participants, using participant group and interlocutor to predict whether a BE or AE expression was produced. There was a group effect in both comparisons, with more AE expressions produced by both ChatGPT and Vicuna than by human participants (ChatGPT: $β = 12.88$, $SE = 0.00$, $z = 37044$, $p < .001$; Vicuna: $β = 5.27$, $SE = 0.64$, $z = 8.30$, $p < .001$; see Fig. 3 bottom right). There was also an interlocutor effect, with more AE expressions when a definition was given by an AE interlocutor than by a BE interlocutor in both ChatGPT and Vicuna compared to in humans (ChatGPT: $β = 2.06$, $SE = 0.00$, $z = 5926$, $p < .001$; Vicuna: $β = 2.10$, $SE = 0.29$, $z = 7.24$, $p < .001$). The interaction was significant in both ChatGPT-human comparison ($β = 2.78$, $SE = 0.00$, $z = 8006$, $p < .001$) and Vicuna-human comparison ($β = 2.82$, $SE = 0.52$, $z = 5.37$, $p < .001$), suggesting that both LLMs were more sensitive to an interlocutor's dialectal background than human participants when producing lexical expressions.